%% file: main.tex
\documentclass[10pt,journal,compsoc]{IEEEtran}
\ifCLASSOPTIONcompsoc
  \usepackage[nocompress]{cite}
\else
  \usepackage{cite}
\fi
\ifCLASSINFOpdf
\else
  
\fi
\usepackage{subfig}
\usepackage{color}
\usepackage{times}
\usepackage{epsfig}
\usepackage{graphicx}
\usepackage{amsmath}
\usepackage{amssymb}

\usepackage{enumitem}
\usepackage{placeins}

\usepackage{url}
\usepackage{booktabs}

\begin{document}

\title{Physics-based Shadow Image Decomposition for Shadow Removal}

\input{definitions}
\def\subFigSzab{\linewidth}
%
%
%
%

\author{ Hieu Le \IEEEcompsocitemizethanks{\IEEEcompsocthanksitem{H. Le is with Amazon Robotics, North Reading, MA 01864. E-mail: hle@cs.stonybrook.edu. This work is done prior to joining Amazon.} \IEEEcompsocthanksitem{D. Samaras is with the Department of Computer Science, Stony Brook University, Stony Brook, NY 11794. E-mail: samaras@cs.stonybrook.edu} } and~Dimitris Samaras 
}



%
%

\markboth{IEEE TRANSACTIONS ON PATTERN ANALYSIS AND MACHINE INTELLIGENCE }%
{Shell \MakeLowercase{\textit{et al.}}: Bare Demo of IEEEtran.cls for Computer Society Journals}
%



\IEEEtitleabstractindextext{%
\begin{abstract}
We propose a novel deep learning method for shadow removal. Inspired by physical models of shadow formation, we use a linear illumination transformation to model the shadow effects in the image that allows the shadow image to be expressed as a combination of the shadow-free image, the shadow parameters, and a matte layer.  We use two deep networks, namely SP-Net and M-Net, to predict the shadow parameters and the shadow matte respectively. This system allows us to remove the shadow effects  from images. We then employ an inpainting network, I-Net, to further refine the results. We train and test our framework on the most challenging shadow removal dataset (ISTD). Our method improves the state-of-the-art in terms of mean absolute error (MAE) for the shadow area by 20\%. Furthermore, this decomposition allows us to formulate a patch-based weakly-supervised shadow removal method. This model can be trained without any shadow- free images (that are cumbersome to acquire) and achieves competitive shadow removal results compared to state-of-the-art methods that are trained with fully paired shadow and shadow-free images. Last, we introduce SBU-Timelapse, a video shadow removal dataset for evaluating shadow removal methods.

\end{abstract}

}

\maketitle

\IEEEdisplaynontitleabstractindextext

%
\IEEEpeerreviewmaketitle

\input{Sec_1_intro.tex}

\input{Sec_2_related_works.tex}
\input{Sec_3_shadow_model.tex}

\input{Sec_4_framework.tex}

\input{Sec_5_weaklysupervised}
\input{Sec_6_Exp}
\input{Sec_7_Conclusion.tex}

\ifCLASSOPTIONcompsoc
  \section*{Acknowledgments}
 
\else
  \section*{Acknowledgment}
\fi

This work was partially supported by the NSF EarthCube program (Award 1740595), the National Neographic/Microsoft AI for Earth program, the Partner University Fund, the SUNY2020 Infrastructure Transportation Security Center, and a gift from Adobe. Computational support provided by the Institute for Advanced Computational Science and a GPU donation from NVIDIA.
We thank Tomas Vicente, Kumara Kahatapitiya, and Cristina Mata for assistance with the prior ICCV 2019 and ECCV 2020 manuscripts.

\ifCLASSOPTIONcaptionsoff
  \newpage
\fi



%

\bibliography{shortstrings,egbib}
\bibliographystyle{IEEEtran}

\begin{IEEEbiography}[{\includegraphics[width=1in,height=1.25in,clip,keepaspectratio]{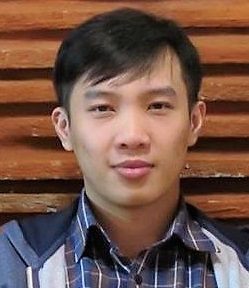}}]{Hieu Le}
received his B.S. degree in computer science from the Ho Chi Minh City University of Science in 2012, and his Ph.D. degree in computer science from the Stony Brook University in 2020. His research interest includes illumination modelling for image analysis and generation. 
\end{IEEEbiography}

\begin{IEEEbiography}[{\includegraphics[width=1in,height=1.25in,clip,keepaspectratio]{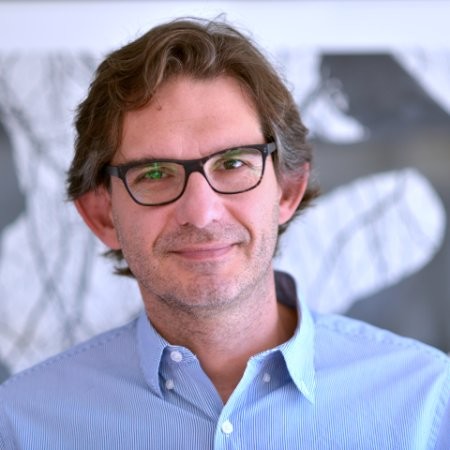}}]{Dimitris Samaras}
received the diploma degree in computer science and engineering from the University of Patras, in 1992, the MSc degree from Northeastern University, in 1994, and the PhD degree from the University of Pennsylvania, in 2001. He is an associate professor of computer science with Stony Brook University. He specializes in illumination modeling and estimation for recognition and graphics,
and biomedical image analysis. He is Program Chair of CVPR 2022.
\end{IEEEbiography}

\end{document}

%% file: definitions.tex
\def\mA{\mathcal{A}}
\def\mB{\mathcal{B}}
\def\mC{\mathcal{C}}
\def\mD{\mathcal{D}}
\def\mE{\mathcal{E}}
\def\mF{\mathcal{F}}
\def\mG{\mathcal{G}}
\def\mH{\mathcal{H}}
\def\mI{\mathcal{I}}
\def\mJ{\mathcal{J}}
\def\mK{\mathcal{K}}
\def\mL{\mathcal{L}}
\def\mM{\mathcal{M}}
\def\mN{\mathcal{N}}
\def\mO{\mathcal{O}}
\def\mP{\mathcal{P}}
\def\mQ{\mathcal{Q}}
\def\mR{\mathcal{R}}
\def\mS{\mathcal{S}}
\def\mT{\mathcal{T}}
\def\mU{\mathcal{U}}
\def\mV{\mathcal{V}}
\def\mW{\mathcal{W}}
\def\mX{\mathcal{X}}
\def\mY{\mathcal{Y}}
\def\mZ{\mathcal{Z}}

\def\1n{\mathbf{1}_n}
\def\0{\mathbf{0}}
\def\1{\mathbf{1}}

\def\A{{\bf A}}
\def\B{{\bf B}}
\def\C{{\bf C}}
\def\D{{\bf D}}
\def\E{{\bf E}}
\def\F{{\bf F}}
\def\G{{\bf G}}
\def\H{{\bf H}}
\def\I{{\bf I}}
\def\J{{\bf J}}
\def\K{{\bf K}}
\def\L{{\bf L}}
\def\M{{\bf M}}
\def\N{{\bf N}}
\def\O{{\bf O}}
\def\P{{\bf P}}
\def\Q{{\bf Q}}
\def\R{{\bf R}}
\def\S{{\bf S}}
\def\T{{\bf T}}
\def\U{{\bf U}}
\def\V{{\bf V}}
\def\W{{\bf W}}
\def\X{{\bf X}}
\def\Y{{\bf Y}}
\def\Z{{\bf Z}}

\def\a{{\bf a}}
\def\b{{\bf b}}
\def\c{{\bf c}}
\def\d{{\bf d}}
\def\e{{\bf e}}
\def\f{{\bf f}}
\def\g{{\bf g}}
\def\h{{\bf h}}
\def\i{{\bf i}}
\def\j{{\bf j}}
\def\k{{\bf k}}
\def\l{{\bf l}}
\def\m{{\bf m}}
\def\n{{\bf n}}
\def\o{{\bf o}}
\def\p{{\bf p}}
\def\q{{\bf q}}
\def\r{{\bf r}}
\def\s{{\bf s}}
\def\t{{\bf t}}
\def\u{{\bf u}}
\def\v{{\bf v}}
\def\w{{\bf w}}
\def\x{{\bf x}}
\def\y{{\bf y}}
\def\z{{\bf z}}

\def\balpha{\mbox{\boldmath{$\alpha$}}}
\def\bbeta{\mbox{\boldmath{$\beta$}}}
\def\bdelta{\mbox{\boldmath{$\delta$}}}
\def\bgamma{\mbox{\boldmath{$\gamma$}}}
\def\blambda{\mbox{\boldmath{$\lambda$}}}
\def\bsigma{\mbox{\boldmath{$\sigma$}}}
\def\btheta{\mbox{\boldmath{$\theta$}}}
\def\bomega{\mbox{\boldmath{$\omega$}}}
\def\bxi{\mbox{\boldmath{$\xi$}}}
\def\bnu{\mbox{\boldmath{$\nu$}}}                                  
\def\bphi{\mbox{\boldmath{$\phi$}}}

\def\bDelta{\mbox{\boldmath{$\Delta$}}}
\def\bOmega{\mbox{\boldmath{$\Omega$}}}
\def\bPhi{\mbox{\boldmath{$\Phi$}}}
\def\bLambda{\mbox{\boldmath{$\Lambda$}}}
\def\bSigma{\mbox{\boldmath{$\Sigma$}}}
\def\bGamma{\mbox{\boldmath{$\Gamma$}}}

\newcommand{\myminimum}[1]{\mathop{\textrm{minimum}}_{#1}}
\newcommand{\mymaximum}[1]{\mathop{\textrm{maximum}}_{#1}}    
\newcommand{\mymean}[1]{\mathop{\textrm{mean}}_{#1}}
\newcommand{\myvar}[1]{\mathop{\textrm{Variance}}_{#1}}
\newcommand{\mymin}[1]{\mathop{\textrm{minimize}}_{#1}}
\newcommand{\mymax}[1]{\mathop{\textrm{maximize}}_{#1}}
\newcommand{\mymins}[1]{\mathop{\textrm{min.}}_{#1}}
\newcommand{\mymaxs}[1]{\mathop{\textrm{max.}}_{#1}}  
\newcommand{\myargmin}[1]{\mathop{\textrm{argmin}}_{#1}} 
\newcommand{\myargmax}[1]{\mathop{\textrm{argmax}}_{#1}} 
\newcommand{\myst}{\textrm{s.t. }}

\newcommand{\denselist}{\itemsep -1pt}
\newcommand{\sparselist}{\itemsep 1pt}

\newcommand{\cyan}[1]{\textcolor{cyan}{#1}}
\newcommand{\red}[1]{\textcolor{red}{#1}}  
\newcommand{\blue}[1]{\textcolor{blue}{#1}}
\newcommand{\magenta}[1]{\textcolor{magenta}{#1}}
\newcommand{\pink}[1]{\textcolor{pink}{#1}}
\newcommand{\green}[1]{\textcolor{green}{#1}} 
\newcommand{\gray}[1]{\textcolor{gray}{#1}}    
\newcommand{\mygreen}[1]{\textcolor{mygreen}{#1}}    
\newcommand{\purple}[1]{\textcolor{purple}{#1}}

\newcommand{\mtodo}[1]{{\color{red}$\blacksquare$\textbf{[TODO: #1]}}}
\newcommand{\myheading}[1]{\vspace{1ex}\noindent \textbf{#1}}

\def\changemargin#1#2{\list{}{\rightmargin#2\leftmargin#1}\item[]}
\let\endchangemargin=\endlist
                                               
\newcommand{\cm}[1]{}

\def\xbi{\overline{\x}_i}
\def\wbi{\overline{\w}_{(i)}}
\def\wb{\overline{\w}}
\def\Ib{\overline{\I}}
\def\invC{\C^{-1}}
\def\invCi{\C_{(i)}^{-1}}
\def\ab{\overline{\balpha}}
\def\abi{\overline{\balpha}_{(i)}}
\def\Kb{\overline{\K}}
\def\Xb{\overline{\X}}
\def\kbi{\overline{\k}_{i}}
\def\Kzz{\K_{\z\z}}
\def\Kzx{\K_{\z\x}}
\def\Xsub{\X_{sub}}
\def\ssub{\s_{sub}}
\def\wbsub{\overline{\w}_{sub}}
\def\dsub{\d_{sub}}
\def\invCsub{\C^{-1}_{sub}}
\def\etal{\emph{et al}.}
\def\etals{\emph{et al}. }
\def\DS{\textcolor{red}}
\newcommand{\norm}[1]{\left\lVert#1\right\rVert}

%% file: Sec_1_intro.tex
\IEEEraisesectionheading{\section{Introduction}\label{sec:introduction}}
Shadows are present in most natural images. They form as a result of complex physical interactions between light sources, geometry, and materials of the objects in the scene.
Analyzing shadows \cite{m_Le-etal-ECCV18,hu2021temporal} gives cues about the physical properties of objects \cite{Okabe09}, illumination conditions \cite{Lalonde10,Panagopoulos09,Panagopoulos13}, scene geometry \cite{karras2018progressive,Sunkavalli2010VisibilitySU}, and  object motions\cite{Zhu2012ObjectbasedCA,Shadow_tracking}. 
Realistic shadow manipulation is an integral part of media editing \cite{Chuang2003, Wang2020PeopleAS} and can greatly improve performance on various computer vision tasks  \cite{Mller2019BrightnessCA,Su2016ShadowDA,Zhang2019ImprovingSS,Surkutlawar2013ShadowSU,DasDocIIW20,SagnikKeICCV2019,Le_2019_CVPR_Workshops,LeICCV2017,Le2016GeodesicDH,JingyiICCV21}. Therefore, methods for detecting and removing shadows have been extensively studied in computer vision. They are getting increased attention in recent years, following the rapid development of deep learning and increasing interest in image generation \cite{CycleGAN2017,karras2018progressive} and augmented virtual reality \cite{Liu2020ARShadowGANSG}. 

\begin{figure}[t]
    \centering
    \includegraphics[width=0.5\textwidth]{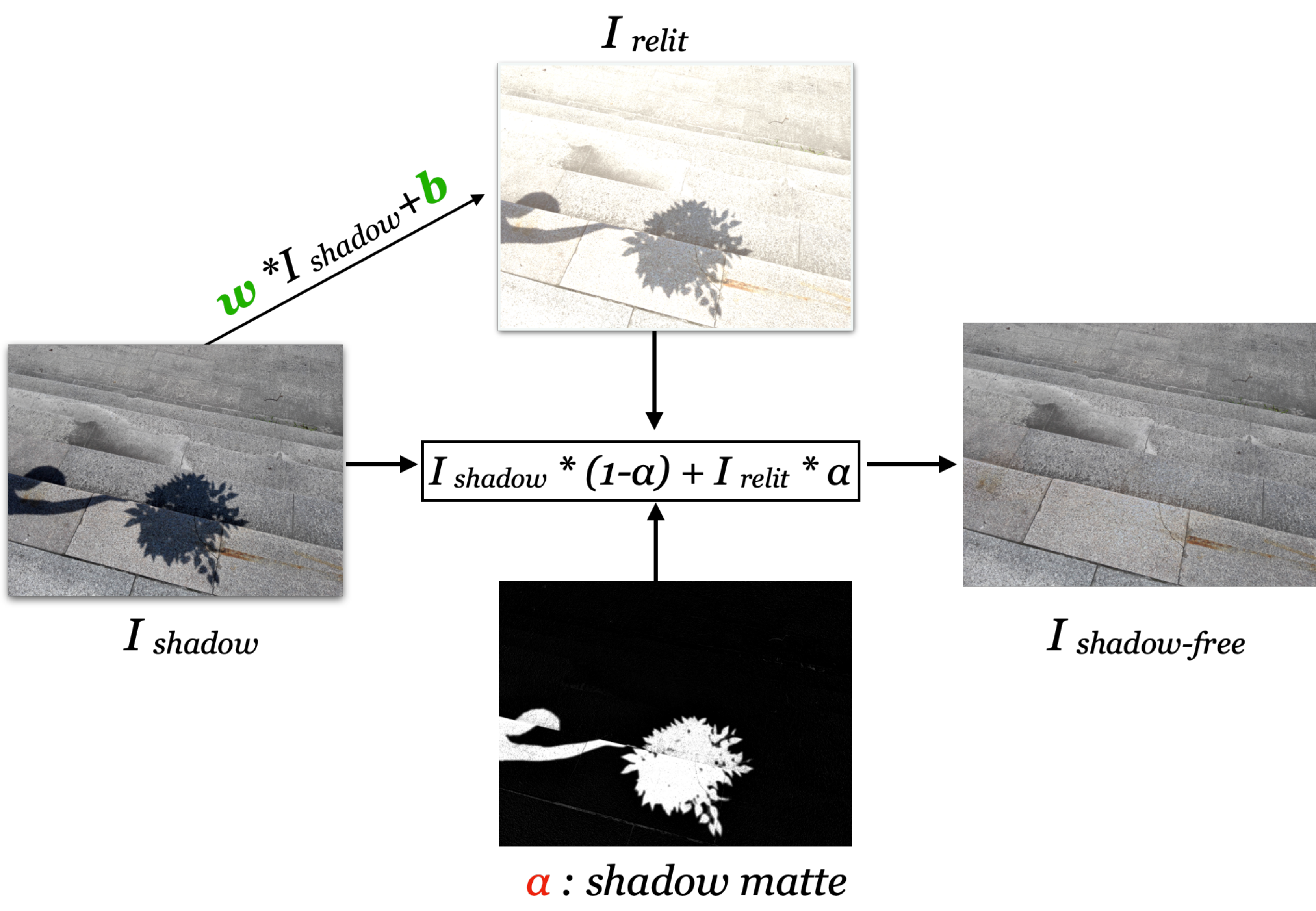}
    \caption{\textbf{Shadow Removal via Shadow Image Decomposition.} A shadow-free image $I_{\textrm{shadow-free}}$ can be expressed in terms of a shadow image $I_{\textrm{shadow}}$, a relit image $I_{\textrm{relit}}$ and a shadow matte $\alpha$. The relit image is a linear transformation of the shadow image. The two unknown factors of this system are the shadow parameters $(w,b)$ and the shadow matte layer $\alpha$. We use two deep networks to estimate these two unknown factors. 
    }
    \label{fig:Teaser}
\end{figure}



Early shadow removal work was based on  physical shadow models \cite{Barrow1978}. A common approach is to formulate the shadow removal problem using an image formation model, in which the image is expressed in terms of material properties and a light source-occluder system that casts shadows. Hence, a shadow-free image can be obtained by estimating the parameters of the source-occluder system and then reversing the shadow effects on the image~\cite{Finlayson06,huang11,guoPami,Shor08}. These methods relight the shadows in a physically plausible manner.
However, estimating the correct solution for such illumination models is non-trivial and requires considerable processing time or user assistance\cite{Zhang15,Chuang2003}.

Alternatively, recently published large-scale datasets \cite{Qu_2017_CVPR,Wang_2018_CVPR,Vicente-etal-ECCV16} allow the use of deep learning methods for shadow removal. In these cases, a network is trained in an end-to-end fashion to map the input shadow image to a shadow-free image. The success of these approaches shows that deep networks can effectively learn transformations that relight shadowed pixels. However, the actual physical properties of shadows are ignored although they contain strong priors to localize and modify shadowed pixels. Without any shadow priors, there is no guarantee that the networks would learn physically plausible transformations. Moreover, there are still well known issues with images generated by deep networks: results tend to be blurry \cite{isola2017image,zhang2016colorful} and/or contain artifacts \cite{odena2016deconvolution}. How to improve the quality of generated images is an active research topic~\cite{karras2018progressive,wang2018pix2pixHD}.

In this work, we propose a novel method for shadow removal that takes advantage of both shadow illumination modelling and deep learning. Following early shadow removal works, we propose to use a simplified physical illumination model to define the mapping between shadow pixels and their shadow-free counterparts. Our main idea is to use deep networks to estimate the parameters for such a physical illumination model. We show that this approach has multiple advantages. The framework removes shadows in a physically plausible manner, avoiding the blurring and undesired artifacts commonly introduced by deep-networks. Moreover, the shadow removal mapping is simpler and easier to constrain since it is strictly defined by the simplified illumination model. The proposed framework achieves the state-of-the-art shadow removal performance via training on paired data, and also can be trained effectively using only unpaired shadow and shadow-free patches, obviating the need of shadow-free image supervision.

Specifically, our proposed illumination model is a linear transformation consisting of a scaling factor and an additive constant - per color channel - for the whole umbra area of the shadow. These scaling factors and additive constants are the parameters of the model. 
 The illumination model plays a key role in our method: with correct parameter estimates, we can use the model to remove shadows from images. Since these parameters are used in a single common linear model to relight the shadows for all shadowed pixels in the input image, this approach ensures global consistency for our shadow removal method. We propose  training a deep network (SP-Net) to automatically estimate the parameters of the shadow model. Through training, SP-Net learns a mapping function from input shadow images to illumination model parameters. 
Furthermore, we use a shadow matting technique \cite{Chuang2003,guoPami,Zhang15} to handle the penumbra area of the shadows. We incorporate our illumination model into an image decomposition formulation~\cite{Porter1984,Chuang2003}, where the shadow-free image is expressed as a combination of the shadow image, the parameters of the shadow model, and a shadow density matte. This image decomposition formulation allows us to reconstruct the shadow-free image, as illustrated in Fig. \ref{fig:Teaser}. The shadow parameters $(w,b)$ represent the transformation from the shadowed pixels to the illuminated pixels. The shadow matte represents the per-pixel linear combination of the relit image and the shadow image, which results to the shadow-free image. Previous work often requires user assistance\cite{Gong16} or solving an optimization system \cite{Levin08} to obtain the shadow mattes. In contrast, we propose to train a second network (M-Net) to accurately predict shadow mattes in a fully automated manner. Lastly, we employ an inpainting network, I-Net, to handle shadow pixels that might not follow our simplified linear illumination model (see Sec.~\ref{Sec:INet}). For example, these cases are due to pixels with saturated colors or shadows with inconsistent intensities across the umbra areas.

In essence, we propose a  method that removes shadows by regressing a set of shadow parameters and a matte layer from the input shadow image. This mapping is simpler than a pixel-wise image-to-image translation and is easier to constrain via regularization of the shadow parameters and matte layer. Thus we can train our proposed shadow removal method even without paired shadow and shadow-free images. In this weakly-supervised setting, we use only shadow and non-shadow patches cropped from the shadow images, which eliminates the need for any shadow-free images. This is an unpaired cross-domain image-to-image translation problem and we approximate this mapping via an adversarial framework.

We trained and tested our model on the ISTD dataset \cite{Wang_2018_CVPR}, which is the largest and most challenging available dataset for shadow removal. Our framework achieves state-of-the-art shadow removal performance on both the fully-supervised and weakly-supervised settings. On the fully-supervised setting, the combination of SP-Net and M-Net alone outperforms the state-of-the-art in shadow removal \cite{hu_pami_2019} by 14\% error reduction in terms of MAE on the shadow areas, from 7.6 to 6.5 MAE. The full framework with SP-Net, M-Net, and the inpainting network I-Net further improves the results by another 8\%, which yields a MAE of 6.0. On the weakly-supervised setting, our weakly-supervised method outperforms the state-of-the-art weakly-supervised shadow removal method\cite{hu_iccv2019mask} by 22\%, reducing the MAE on the shadow areas from 12.4 to 9.7. Note that our method is the first deep-learning approach that can learn a shadow removal transformation without any shadow-free images. 


Last, we introduce a novel challenging testing set to evaluate shadow removal methods based on time-lapse videos. Note that all current shadow removal datasets\cite{Qu_2017_CVPR,Wang_2018_CVPR} are limited to simple shadows, simple scenes, and without the oclluders in the images due to the data acquisition scheme\cite{Qu_2017_CVPR,Wang_2018_CVPR}. With time-lapse videos where the only motions are due to illumination effects, we can obtain paired shadow and non-shadow pixels that move in and out of the shadows when they travel across the scene.
 This method allows us to collect paired shadow data that is impossible to obtain via the standard data acquisition method such as shadows of immobile occluders, self-cast shadows, and shadows on various types of backgrounds and materials. We introduce SBU-Timelapse, a video dataset of 50 clips. Our methods perform better than other methods on this challenging test.






In summary, the contributions of this work are:
\begin{itemize}
    \item We propose a new deep learning approach for shadow removal, grounded by a simplified physical illumination model and an image decomposition formulation.
    
    \item We propose a weakly-supervised training scheme for our method that can be trained without any shadow-free images. 
  
    \item Our proposed approaches achieve state-of-the-art shadow removal results on the ISTD dataset, on both fully-supervised and weakly-supervised settings. 
    
    \item We propose a method to obtain paired shadow data of complex shadows in complex scenes based on time-lapse videos. We introduced SBU-Timelapse, a video shadow removal dataset for evaluating shadow removal methods. 
\end{itemize}

The pre-trained model, shadow removal results, and more details can be found at: \url{https://github.com/cvlab-stonybrook/SID}

%% file: Sec_2_related_works.tex
\begin{figure*}[!t]
 \centering
    \includegraphics[width=\textwidth]{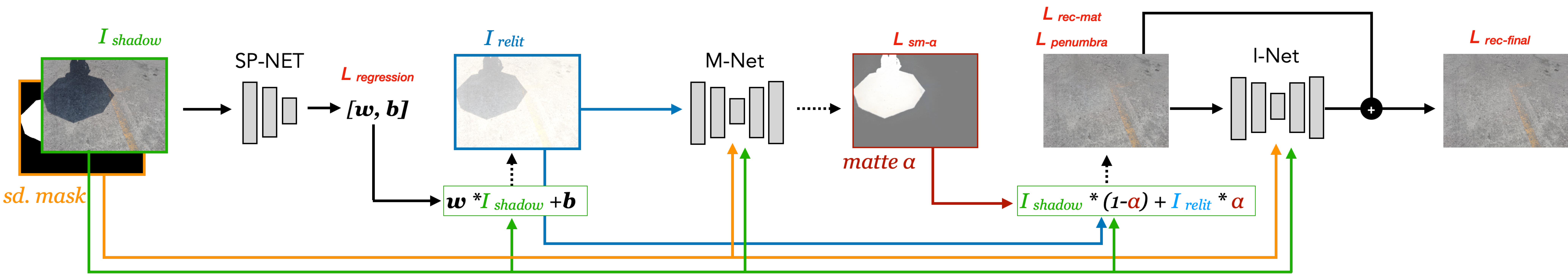}
    \caption{\textbf{Shadow Removal Framework.} The shadow parameter estimator network SP-Net takes as input the shadow image and the shadow mask to predict the shadow parameters $(w,b)$. The relit image  $I^{\textrm{relit}}$ is then computed via Eq. \ref{eq:relit} using the estimated parameters from SP-Net. The relit image, together with the input shadow image and the shadow mask are then input into the shadow matte prediction network M-Net to get the shadow matte layer $\alpha$. The system outputs the shadow-free image via Eq. \ref{eq:decom}, using the shadow image, the relit image, and the shadow matte. SP-Net learns to predict the shadow parameters $(w,b)$, denoted as the regression loss. M-Net learns to minimize the $L_1$ distance between the output of the system and the shadow-free image (reconstruction loss). I-Net takes as input the shadow image, shadow mask, and the shadow-free image computed from SP-Net and M-Net to output a residual image that is used to obtain the final shadow-free image. 
    }
    \label{fig:framework}
\end{figure*}

\section{Related Work}
\label{sec:related}
\textbf{Shadow Illumination Models:}
Early research on shadow removal is motivated by physical modelling of illumination and color \cite{Finlayson06,Finlayson01,Finlayson02,Drew03recoveryof}.
Barrow \& Tenenbaum \cite{Barrow1978} define an intrinsic image algorithm that separates images into the intrinsic components of reflectance and shading.  Guo \etal~\cite{guoPami} simplify this model to represent the relationship between the shadow pixels and shadow-free pixels via a linear system. 
They estimate the unknown factors via pairing shadow and shadow-free regions. Similarly, Shor \& Lischinki \cite{Shor08} propose an illumination model for shadows in which there is an affine relationship between the lit and shadow intensities at a pixel, including 4 unknown parameters. They define two strips of pixels: one in the shadowed area and one in the lit area to estimate their parameters.
Finlayson \etal \cite{Finlayson09} create an illuminant-invariant
image for shadow detection and removal. Their work is based on an insight that the shadowed pixels differ from their lit pixels by a scaling factor. Vicente \etal~\cite{Vicente-etal-PAMI18,vicentesingle}  propose a method for shadow removal where they suggest that the color of the lit region can be transferred to the shadowed region via histogram equalization. 


\textbf{Shadow Matting:} 
Matting, introduced by Porter \& Duff \cite{Porter1984}, is an effective tool to handle soft shadows. However, it is non-trivial to compute the shadow matte from a single image.
Chuang \etal~\cite{Chuang2003} use image matting for shadow editing to transfer the shadows between different scenes. They compute the shadow matte from a sequence of frames in a video captured from a static camera.  
Guo et al. \cite{guoPami} and Zhang \etal~\cite{Zhang15} both use a shadow matte for their shadow removal frameworks, where they estimate the shadow matte via the closed-form solution of Levin \etal~\cite{Levin08}. 

\textbf{Deep-Learning Based Shadow Removal:} 
Recently published large-scale datasets~\cite{Vicente-etal-ECCV16, Wang_2018_CVPR,Qu_2017_CVPR} enable training deep-learning networks for shadow removal.
The Deshadow-Net of Qu \etal~\cite{Qu_2017_CVPR} is trained to remove  shadows in an end-to-end manner.  Their network extracts multi-context features across different layers of a deep network to predict a shadow matte. This shadow matte is different from ours as it contains both the density and color offset of the shadows. The ST-CGAN proposed by Wang \etal~\cite{Wang_2018_CVPR} for both shadow detection and removal is a conditional GAN-based framework \cite{isola2017image} for shadow detection and removal. Their framework is trained to predict the shadow mask and shadow-free image in a unified manner, they use GAN losses to improve performance. Mask-ShadowGAN \cite{Hu_2018_CVPR} is the only deep-learning method that learns shadow removal from just unpaired training data. Their framework is based on the CycleGAN \cite{CycleGAN2017}, which approximates a mapping from a source domain, i.e., shadow-images, to a target domain - shadow-free images. By contrast, our weakly-supervised method can be trained using only shadow-images. 

Inspired by early work, our framework outputs the shadow-free image based on a physically inspired shadow illumination model and a shadow matte. We, however, estimate the parameters of our model and the shadow matte via two deep networks in a fully automated manner.


%% file: Sec_3_shadow_model.tex
\section{Shadow and Image Decomposition Model}
\label{sec:decom_model}
\subsection{Shadow Illumination Model}
Let us begin by describing our shadow illumination model. We aim to find a mapping function $T$ to transform a shadow pixel $I^{\textrm{shadow}}_x$ to its non-shadow counterpart: $I_x^{\textrm{shadow-free}} = T(I^{\textrm{shadow}}_x,w)$ where $w$ are the parameters of the model. The form of $T$ has been studied in depth in  previous work as  discussed in Sec. \ref{sec:related}. 

In this paper, similar to the model of Shor \& Lischinski \cite{Shor08}, we use a linear function to model the relationship between the lit and shadowed pixels. The intensity of a lit pixel is formulated as:
\begin{equation}
\label{eq:shadfree}
 I^{\textrm{shadow-free}}_x(\lambda) =  L^d_x(\lambda)R_x(\lambda) +   L^a_x(\lambda)R_x(\lambda)
\end{equation}
where $ I^{\textrm{shadow-free}}_x(\lambda)$ is the intensity reflected from point $x$ in the scene at wavelength $\lambda$, $L$ and $R$ are the illumination and reflectance respectively, $L^d$ is the direct illumination and $L^a$ is the ambient illumination.

To  cast a shadow on point $x$,  an occluder blocks the direct illumination and a portion of the ambient illumination that would otherwise arrive at $x$. The shadowed intensity at $x$ is:

\begin{equation}
\label{eq:shad}
 I^{\textrm{shadow}}_x(\lambda) =  a_x(\lambda)L^a_x(\lambda)R_x(\lambda)
\end{equation}
where $a_x(\lambda)$ is the attenuation factor indicating the remaining fraction of the ambient illumination that arrives at point $x$ at wavelength $\lambda$.
Note that Shor \& Lischinski further assume that $a_x(\lambda)$ is the same for all wavelengths $\lambda$ to simplify their model. This assumption implies that the environment light has the same color from all directions.

From Eqs.\ref{eq:shadfree} and \ref{eq:shad}, we can express the shadow-free pixel as a linear function of the shadowed pixel:
\begin{equation}
\label{eq:maineq}
 I^{\textrm{shadow-free}}_x(\lambda) =  L^d_x(\lambda)R_x(\lambda) +  a_x(\lambda)^{-1}I^{\textrm{shadow}}_x(\lambda)
\end{equation}

We assume that this linear relation is preserved throughout the color acquisition process of the camera \cite{Finlayson16}. Therefore, we can express the color intensity of the lit pixel $x$ as a linear function of its shadowed value for all pixels in the shadow area:
\begin{equation}
\label{eq:trans}
    I_x^{\textrm{shadow-free}}(k) = w_k \times I_x^{\textrm{shadow}}(k) + b_k
\end{equation}
where $I_x(k)$ represents the value of the pixel $x$ on the image $I$ in  color channel $k$ ($k\in {\textrm{R,G,B color channel}}$), $b_k$ is the response of the camera to  direct illumination, and $w_k$ is responsible for the attenuation factor of the ambient illumination at this pixel in this color channel. We model each color channel independently to account for possibly different spectral characteristics of the material in shadow as well as the sensor.

We further assume that the two vectors $w =[w_R,w_G,w_B]$ and $b=[b_R,b_G,b_B]$ are constant across all pixels $x$ in the umbra area of the shadow. Under this assumption, we can easily estimate the values of $w$ and $b$  given the shadow and shadow-free image using linear regression.  We refer to $(w,b)$ as the \textit{shadow parameters} in the rest of the paper. 

In Sec. \ref{sec:framework}, we show that we can train a deep-network to estimate these vectors from a single image.

\subsection{Shadow Image Decomposition System}
\label{sec:decom}
We incorporate our proposed shadow illumination model into the following well-known image decomposition system \cite{Chuang2003,Porter1984,Smith1996,Wright}. The system models the shadow-free image using the shadow image, the shadow parameter, and the shadow matte. The shadow-free image can be expressed as:

\begin{equation}
    I^{\textrm{shadow-free}} = I^{\textrm{shadow}} \cdot (1-\alpha) + I^{\textrm{relit}}\cdot \alpha
    \label{eq:decom}
\end{equation}
where  $I^{\textrm{shadow}}$ and $I^{\textrm{shadow-free}}$ are the shadow and shadow-free image respectively, $\alpha$ is the matting layer, and $I^{\textrm{relit}}$ is the relit image. We define $\alpha$ and $I^{\textrm{relit}}$ below.

Each pixel $i$ of the relit image  $I^{\textrm{relit}}$  is computed by:

\begin{equation}
    I_i^{\textrm{relit}} = w \cdot I_i^{\textrm{shadow}} +b
    \label{eq:relit}
\end{equation}
which is the shadow image  transformed by the illumination model of Eq. \ref{eq:trans}. This transformation maps the shadowed pixels to their shadow-free values.

The matting layer $\alpha$ represents the per-pixel coefficients of the linear combination of the relit image and the input shadow image that results into the shadow-free image. Ideally, the value of $\alpha$ should be 1 at the non-shadow area and 0 at the umbra of the shadow area. For the pixels in the penumbra of the shadow, the value of $\alpha$ gradually changes near the shadow boundary.

The value of $\alpha$ at pixel $i$ based on  the shadow image, shadow-free image, and relit image, follows from Eq. \ref{eq:decom} :
\begin{equation}
\label{eq:alpha}
\alpha_i=\frac{{I_i}^{\textrm{shadow-free}}-{I_i}^{\textrm{shadow}}}{{I_i}^{\textrm{relit}}-{I_i}^{\textrm{shadow}}}
\end{equation}

We use the image decomposition of Eq. \ref{eq:decom} for our shadow removal framework. The  unknown factors are the shadow parameters $(w,b)$ and the shadow matte $\alpha$. We present our method that uses two deep networks, SP-Net and M-Net, to predict these two factors in the following section. Sec. \ref{sec:framework} presents our fully-supervised framework trained using paired shadow and shadow-free images. In Sec. \ref{sec:weakly}, we present our weakly-supervised framework trained using only shadow images.


%% file: Sec_4_framework.tex
\section{Fully-Supervised Shadow Removal Framework}
\label{sec:framework}
Our shadow removal framework is based on the shadow image decomposition model in Eq. \ref{eq:decom}. The unknowns are the set of shadow parameters $(w,b)$ and the shadow matte $\alpha$. We train two deep networks, namely, SP-Net, and M-Net to predict the shadow parameters and the shadow matte respectively. We then train an inpainting network, I-Net, to handle cases that might not follow the linear illumination model.

Fig. \ref{fig:framework} summarizes our fully-supervised shadow removal framework. The shadow parameter estimator network SP-Net takes as input the shadow image and the shadow mask to predict the shadow parameters $(w,b)$. The relit image  $I^{\textrm{relit}}$ is then computed via Eq. \ref{eq:relit} with the estimated parameters from SP-Net. The relit image, together with the input shadow image and the shadow mask is then input into the shadow matte prediction network M-Net to get the shadow matte $\alpha$. The system outputs a shadow-free image via Eq. \ref{eq:decom}. I-Net takes as input the shadow image, shadow mask, and the shadow-free image computed from SP-Net and M-Net to output a residual image that is used to obtain the final shadow-free image.

\subsection{Shadow Parameter Estimator Network}
\label{sec:SP-Net}

In order to recover the illuminated intensity at the shadowed pixel, we need to estimate the parameters of the linear model in Eq. \ref{eq:trans}. Previous work has proposed different methods to estimate the parameters of a shadow illumination model \cite{Shor08,Gong16,guoPami,Finlayson02,Finlayson09,Drew03recoveryof}. 
In this paper, we train SP-Net, a deep network model, to directly predict the shadow parameters from the input shadow image.

To train SP-Net, we first generate training data.
Given a training pair of a shadow image and a shadow-free image, we estimate the parameters of our linear illumination model using a least squares method \cite{cook1986,Sun2010OptimizationTA}. For each shadow image, we first erode the shadow mask by 5 pixels in order to define a region that does not contain the partially shadowed (penumbra) pixels. Mapping these shadow pixel values to the corresponding values in the shadow-free image, gives us a linear regression system, from which we calculate $w$ and $b$.  
We further bound the search space of  $w$ to [1,3]. The minimum value of $w=1$ ensures that the transformation always scales up the values of the shadowed pixels while the upper bound is determined based on the training set. 
We use the Trust Region Algorithm \cite{Sun2010OptimizationTA} to solve this linear system, which is particularly suitable for solving optimization problems with bounds. The Trust Region Method defines a region around the current best solution such that a model could approximate the original function within this region. The trusted region is then updated based on the new best solution.

We compute parameters for each of  the three RGB color channels  and then combine the learned coefficients to form a  6-element vector. This vector is used as the targeted output to train SP-Net. The input for SP-Net is the input shadow image and the associated shadow mask. We train SP-Net to minimize the two losses: 1) the $L_1$ distance between the output of the network and these computed shadow parameters, marked as ``$\mathcal{L}_\textrm{regression}$'' in Fig. \ref{fig:framework}, and 2) the final $L_1$ image reconstruction loss, marked as ``$\mathcal{L}_\textrm{rec-final}$''. 
Notice that while we use the ground truth shadow mask for training, during testing we estimate shadow masks using the shadow detection network proposed by Zhu \etal \cite{zhu18b}.

\subsection{Shadow Matte Prediction Network}
\label{Sec:MNet}

\def\subboxsize{0.16\textwidth}
\begin{figure*}[th!]
    \centering
    \includegraphics[width=\textwidth]{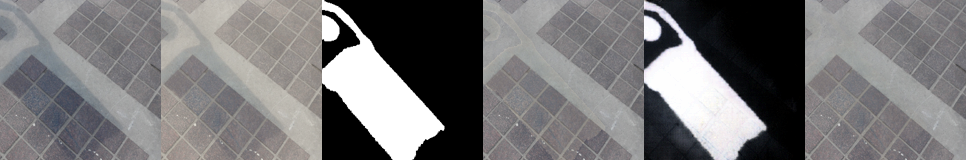}\\
    \makebox[\subboxsize]{Input}
    \makebox[\subboxsize]{Relit}
    \makebox[\subboxsize]{Shad. Mask}
    \makebox[\subboxsize]{Using S.Mask}
    \makebox[\subboxsize]{Shad. Matte}
    \makebox[\subboxsize]{Using S.Matte}

    \caption{\textbf{A comparison of the ground truth shadow mask and our shadow matte.} From  left to right: The input image, the relit image computed from the parameters estimated via SP-Net, the ground truth shadow mask, the final results when we use the shadow mask, the shadow matte computed using our M-Net, and the final shadow-free image  when we use the shadow matte to combine the input and relit image. The matting layer handles the soft shadow and does not generate visible boundaries in the final result. \em{(Please view in magnification on a digital device to see the difference more clearly.)}}
    \label{fig:matte}
\end{figure*}

Our linear illumination model (Eq. \ref{eq:trans}) can relight the pixels in the umbra area (fully shadowed). The shadowed pixels in the penumbra (partially shadowed) region are more challenging as the illumination changes gradually across the shadow boundary \cite{huang11}. A binary shadow mask cannot model this gradual change. Thus, using a binary mask within the decomposition model in Eq. \ref{eq:decom} will generate an image with visible boundary artifacts. A solution for this is shadow matting where the soft shadow effects are expressed via the values of a blending layer.

In this paper, we train a deep network, M-Net, to predict this matting layer. In order to train M-Net, we use Eq. \ref{eq:decom} to compute the output of our framework, where the shadow matte is the output of M-Net. Then the first loss function that drives the training of M-Net is an $L_1$ distance between the  output image at this stage, i.e.,  obtained using only SP-Net and M-Net, and the ground truth training shadow-free image, marked as ``$\mathcal{L}_\textrm{rec-mat}$'' in Fig. \ref{fig:framework}. To simply force M-Net to focus on penumbra areas, we use an additional $L_1$ reconstruction loss function applied for these areas only, denoted as ``$\mathcal{L}_\textrm{penumbra}$''. We define the penumbra areas of a shadow as the combination of two areas alongside the shadow boundary, denoted as $\mM_{in}$ and $\mM_{out}$ - see Fig.\ref{fig:boundary}. $\mM_{out}$ is the area right outside the boundary, computed by subtracting the shadow mask, $\mM$, from its dilated version $\mM_{dilated}$. The inside area $\mM_{in}$ is computed similarly by subtracting an eroded shadow mask from the shadow mask.

\def\subboxsize{0.155\textwidth}

\begin{figure}[b]
	\centering
  \includegraphics[width=0.45\textwidth]{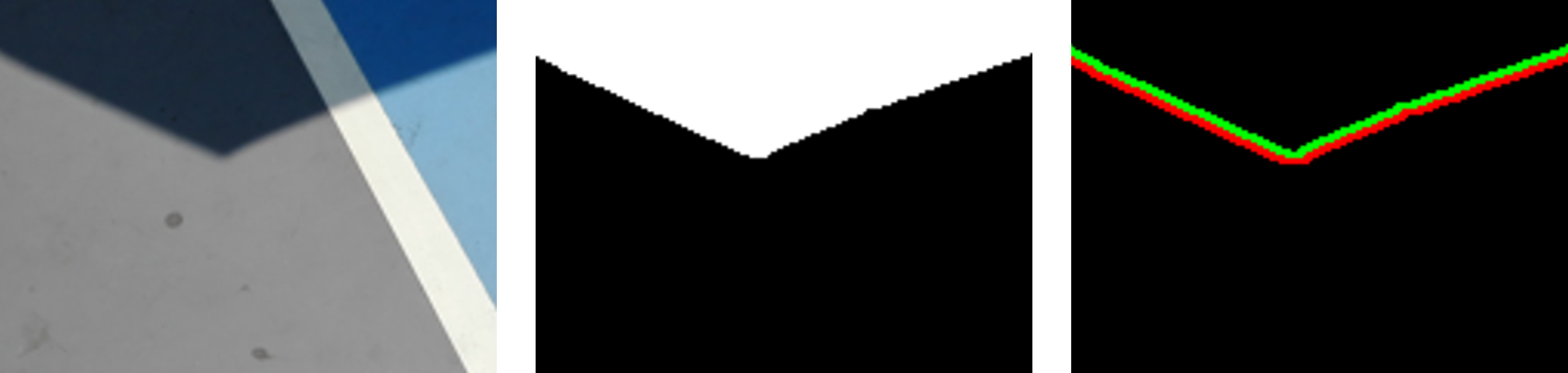}
    \makebox[\subboxsize]{Input Image}
    \makebox[\subboxsize]{Shadow Mask}
    \makebox[\subboxsize]{$\mM_{in}$ \& $\mM_{out}$}

  \caption{\textbf{The penumbra area of the shadow.} We define two areas alongside the shadow boundary, denoted as $\mM_{in}$ (shown in green) and $\mM_{out}$ (shown in red). These two areas roughly define a small region surrounding the shadow boundary, which can be considered as the penumbra area of the shadow. }
  \label{fig:boundary}

\end{figure}

Moreover, since the shadow effects are assumed to vary smoothly across the shadow boundaries, we enforce an $L_1$ smoothness loss on the spatial gradients of the matte layer, $\alpha$. This smoothness loss $\mathcal{L}_{sm-\alpha} $ also prevents M-Net from producing undesired artifacts since it enforces local uniformity. This loss is:

\begin{equation}
\mathcal{L}_{sm-\alpha} =|\nabla{\alpha}| 
\end{equation}

Fig. \ref{fig:matte} illustrates the effectiveness of our shadow matting technique. We show in the figure two shadow removal results which are computed using a ground-truth shadow mask and a shadow matte respectively. This shadow matte is computed by our model. One can see that using the binary shadow mask to form the shadow-free image  creates visible boundary artifacts as it ignores the penumbra. The shadow matte  from our model captures well the soft shadow and generates an image without shadow boundary artifacts.



\subsection{Inpainting Network}
\label{Sec:INet}

Our model removes shadows in umbra areas via the simplified physical illumination model. However, there are shadow pixels that might not follow our proposed illumination model due to various reasons:

\begin{itemize}
    \item Pixels with low or high intensities might not follow our illumination model due to  color saturation, i.e., when shadows are cast on very ``dark'' pixels or relighting very ``bright'' pixels.  

    \item Our assumptions are strict. A single set of shadow parameters might not represent all shadow pixels. For example, shadow intensity can be inconsistent across the shadowed region. Shadow pixels that are closer to the occluding object often have darker shadows due to more ambient illumination being blocked. 
    
    \item There are errors in data acquisition which might lead to incorrect estimates of the shadow parameters, for example, the inconsistent ambient lighting between the shadow and shadow-free images.

\end{itemize}

To mitigate these issues, we train an inpainting network, I-Net, to output a residual layer to better reconstruct shadow-free images. Previous deep-learning works also remove shadows via pixel-wise residual layers \cite{Qu_2017_CVPR,hu_pami_2019}. However, they typically estimate a residual layer to remove the shadows directly from the input shadow image while we estimate a residual layer to further refine the shadow removal result of the proposed SP-Net and M-Net. %
Fig.\ref{fig:inpainting} visualizes the effect of our inpainting network. The first row shows an example where the shadow intensity is inconsistent across the shadowed region. The second row shows an example where pixels with dark colors might not follow our illumination model. The third row shows an example where our SP-Net does not estimate correctly the shadow parameters. I-Net mitigates these issues, as visualized in the last two columns.  The loss function that drives the training of I-Net is a reconstruction loss penalizing an $L_1$ distance between output image and ground truth training shadow-free image, denoted as $\mathcal{L}_{rec-final}$ in Fig.\ref{fig:framework}.

\def\subboxsize{0.19\textwidth}
\def\cwidth{0.95\textwidth}
\begin{figure*}[th]
    \centering
    \includegraphics[width=\cwidth]{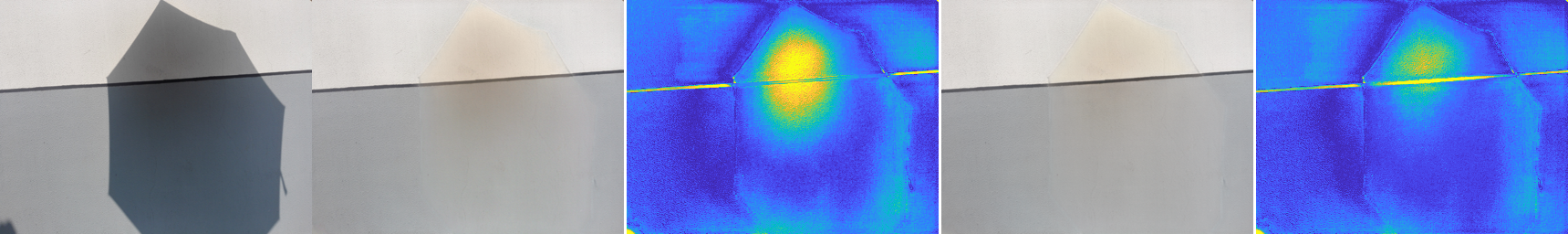}\\
    \includegraphics[width=\cwidth]{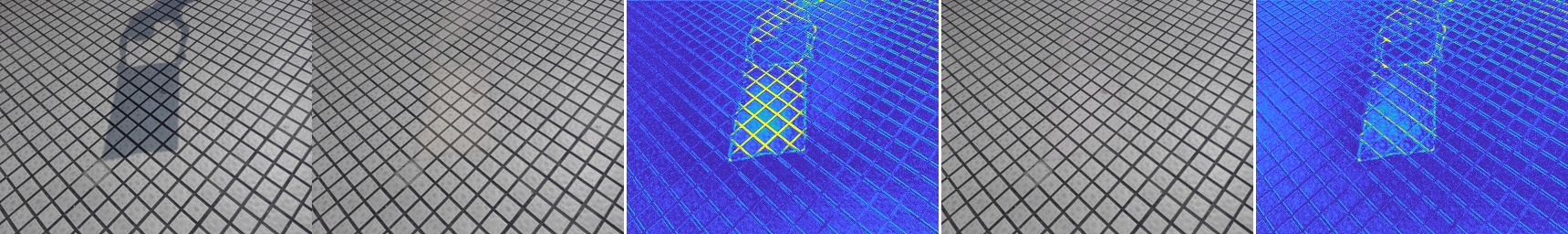}\\
    \includegraphics[width=\cwidth]{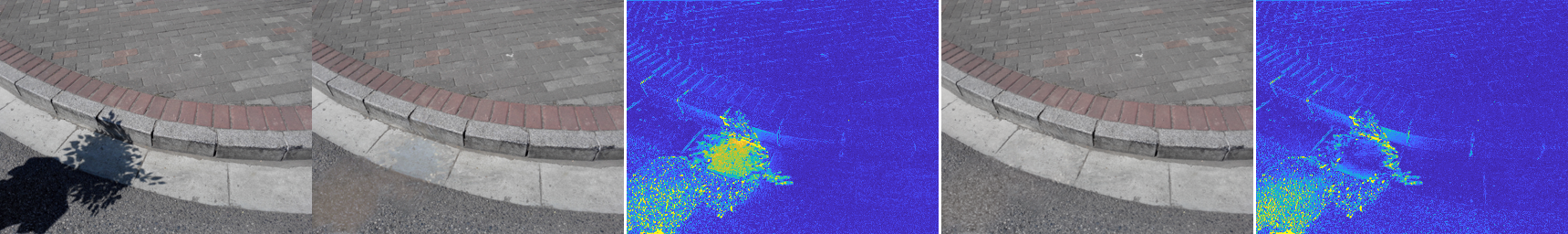}\\
            
    \makebox[\subboxsize]{Input}
    \makebox[\subboxsize]{SP+M-Net}
    \makebox[\subboxsize]{SP+M Error Map}
    \makebox[\subboxsize]{SP+M+I-Net}
    \makebox[\subboxsize]{SP+M+I Error Map}

    \caption{\textbf{The effect of the inpainting network, I-Net.} From left to right: The input images, the shadow-free images computed from our SP-Net and M-Net (SP+M-Net), the error heat maps of the SP+M-Net's results, the shadow-free images computed with the addition of the inpainting network I-Net (SP+M+I-Net), and the error heat maps of SP+M+I-Net's results. The inpainting network I-Net partially corrects the colors of the pixels in the umbra areas.}
    \label{fig:inpainting}
\end{figure*}

\subsection{ Objective Function}
Fig.\ref{fig:framework} illustrates the loss functions that drive the training of our framework. The final objective function for training our system aims to minimize a weighted sum of all the losses:

\begin{align*}
\mathcal{L}_{final} =& \lambda_{reg}\mathcal{L}_{regression} +\lambda_{sm}\mathcal{L}_{sm-\alpha} +\lambda_{pen}\mathcal{L}_{penumbra} \\ 
& +\lambda_{rec-mat}\mathcal{L}_{rec-mat}+ \lambda_{rec-final}\mathcal{L}_{rec-final}
\end{align*}


%% file: Sec_5_weaklysupervised.tex
\section{Weakly-supervised Shadow Removal}
\label{sec:weakly}

\begin{figure*}[h]
	\centering
  \includegraphics[width=0.9\textwidth]{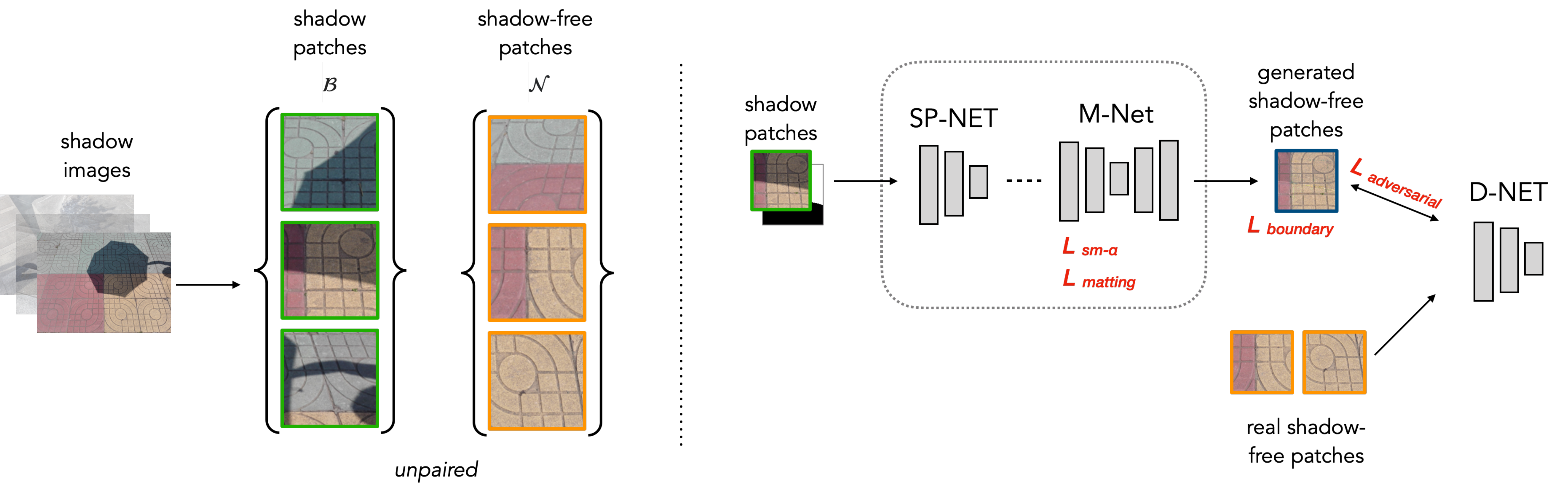}

  \caption{\textbf{Weakly-supervised shadow decomposition.} We train our SP-Net and M-Net together with an adversarial network, D-Net, using only two sets of unpaired shadow and shadow free patches. These patches are cropped directly from the shadow images (left panel). SP-Net and M-Net predict the shadow parameters $(w,b)$ and the matte layer $\alpha$ respectively to jointly remove the shadow. D-Net is the critic function distinguishing between the generated image patches and the real shadow-free patches. The only supervision signal is the set of shadow-free patches, which can be easily obtained using the shadow masks. The four losses guiding this training are the matting loss, smoothness loss, boundary loss, and adversarial loss.}
  \label{fig:weakly_framework}
\end{figure*}

\def\subfig{0.9\textwidth}
\def\subboxsize{0.15\textwidth}
\begin{figure*}[h]
 \centering

  \includegraphics[width=\subfig]{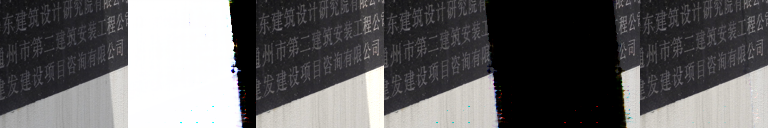}
    \makebox[\subboxsize]{$I^{sd}$}
    \makebox[\subboxsize]{$\alpha$}
    \makebox[\subboxsize]{$I^{relit}$}
    \makebox[\subboxsize]{$I^{relit}*\alpha$}
    \makebox[\subboxsize]{$I^{sd}*(1-\alpha)$ }
    \makebox[\subboxsize]{$I^{output}$}\\
     \caption{\textbf{Weakly-supervised shadow image decomposition.} With only shadow mask supervision, our method automatically learns to decompose the shadow effect in the input image patch $I^{sd}$ into a matte layer $\alpha$ and a relit image $I^{relit}$. The matte layer $\alpha$ combines $I^{sd}$  and $I^{relit}$ to obtain a shadow-free image patch $I^{output}$ via Eq. (\ref{eq:decom}).
    }
 
    \label{fig:decompose}

\end{figure*}

The principle of our shadow removal method is to decompose a shadow image into its intrinsic components including a set of shadow parameters which explain the shadow effects in the umbra areas, and a matte layer that models the soft shadows. In this section, we show that such an image decomposition method is easy to constrain and models closely a shadow removal transformation. Thus, it can even be trained with a much weaker form of supervision signal rather than paired shadow and shadow-free images. In this section, we use only two sets of unpaired patches, shadow and shadow-free, to train our shadow decomposition framework. Since these patches are directly cropped from the shadow images, this method can be trained without any shadow-free images.

To obtain shadow and shadow-free patches, we crop the shadow images into small overlapping patches of size $n\times n$ with a step size of $m$. Based on the shadow masks, we group these patches into three sets: a non-shadow set ($\mN$) containing patches having no shadow pixels, a shadow-boundary set ($\mB$) containing patches lying on the shadow boundaries, and a full-shadow set ($\mF$) containing patches where all pixels are in shadow.
With small enough patch size $n$ and  step size $m$, we can obtain enough training patches in each set. With this training set, we train a shadow removal system to learn a mapping  from patches in the shadow-boundary set $\mB$ to patches in the non-shadow set $\mN$.

Fig. \ref{fig:weakly_framework} summarizes our weakly-supervised training scheme, which consists of three networks: SP-Net, M-Net, and D-Net. SP-Net and M-Net predict the shadow parameters $(w,b)$ and the matte layer $\alpha$ respectively to jointly remove shadows. D-Net is the critic distinguishing between the generated image patches and the real shadow-free patches. With SP-Net and M-Net being the generators and D-Net being the discriminator, the three networks form an adversarial training framework where the main source of training signal is the set of shadow-free patches.

In theory, as D-Net is trained to distinguish patches containing shadow boundaries from patches without any shadows, a natural solution to fool D-Net is to remove the shadows in the input shadow patches to make them indistinguishable from shadow-free patches. However, such an adversarial signal from D-Net alone often cannot guide the generators, (SP-Net and M-Net) to actually remove shadows. The parameter search space is very large and the mapping is extremely under-constrained. In practice, we observe that without any constraints, SP-Net tends to output consistently high values of $(w,b)$ as they would directly increase the overall brightness of the image patches, and Matte-Net tends to introduce artifacts similar to visual patterns frequently appearing in the non-shadow areas.
Thus, our main intuition is to constrain this framework with physical shadow properties to force the networks to only transform the input images in a manner consistent with shadow removal. 

First, we bound the search space of SP-Net and M-Net to the appropriate value ranges that correspond to shadow removal. The search space for the shadow parameters is defined similarly to Sec. \ref{sec:SP-Net}. M-Net estimates a blending layer $\alpha$ that combines the shadow image patch and the relit image patch into a shadow-free image patch via Eq. \ref{eq:decom}. Thus, we map the output of M-Net to [0,1] as $\alpha$ is being used as a matting layer and constrain the value of $\alpha_i$ as follows:
\begin{itemize}
    \item If \textit{i} indicates a non-shadow pixel, we enforce $\alpha_i=0$ so that the value of the output pixel $I_i^{\textrm{output}}$ equals its value in the input image $I_i^{\textrm{shadow}}$.
    \item If \textit{i} indicates a pixel in the umbra areas of the shadows, we enforce $\alpha_i=1$ so that the value of the output pixel $I_i^{\textrm{output}}$ equals its relit value $I_i^{\textrm{relit}}$.
    \item We do not control the value of $\alpha$ in the penumbra areas of the shadows; they are learnt though training.
\end{itemize}

\noindent where the umbra, non-shadow or penumbra areas can be roughly specified using the shadow mask. The above constraints are implemented as the matting loss $\mathcal{L}_{mat-\alpha} $ computed by the following formula for every pixel $i$:

\begin{equation}
\mathcal{L}_{mat-\alpha} = \sum_{i\in (\mM - \mM_{in})} |\alpha_i-1| +  \sum_{i\notin \mM_{dilated}} |\alpha_i|
\end{equation}


Moreover, we enforce an $L1$ smoothness loss on the spatial gradients of the matte layer, $\alpha$:

\begin{equation}
\mathcal{L}_{sm-\alpha} =|\nabla{\alpha}| 
\end{equation}
These constraints play a vital part in our weakly-supervised framework, as we show in the ablation study in Sec.\ref{sec:ablation}.

Then, given a set of estimated parameters $(w,b)$ and a matte layer $\alpha$, we obtain an output image $I^{\textrm{output}}$ via the image decomposition formula (Eq.\ref{eq:decom}). We penalize the $L1$ difference between the average intensity of pixels lying right outside and inside the shadow boundary,  $\mM_{in}$ and $\mM_{out}$ respectively. This shadow boundary loss $\mathcal{L}_{bd}$ is computed by:

\begin{equation}
\mathcal{L}_{bd} = \left|  \frac{\sum_{i\in \mM_{in}} I_i^{output}}{\sum_{i\in \mM_{in}} }  -  \frac{\sum_{i\in \mM_{out}}  I_i^{output}}{\sum_{i\in \mM_{out}}}\right|
\end{equation}

Last, we compute the adversarial loss with the feedback from D-Net: 
\begin{equation}
\mathcal{L}_{GAN} =\log (1-D(I^{output})) 
\end{equation}
where $D(\cdot)$ denotes the output of D-Net.

The final objective function to train Param-Net and Matte-Net is to minimize a weighted sum of the above losses:
\begin{equation}
\mathcal{L}_{final} = \lambda_{sm}\mathcal{L}_{sm-\alpha} + \lambda_{mat}\mathcal{L}_{mat-\alpha} + \lambda_{bd}\mathcal{L}_{bd}+ \lambda_{adv}\mathcal{L}_{GAN}
\end{equation}

By using all the proposed losses together, our method is able to automatically extract a set of shadow parameters and an $\alpha$ layer from an input image patch. Fig. \ref{fig:decompose} visualizes the components extracted from our framework for an input patch. 

Based on our patch-based model, we can estimate a set of shadow parameters and a matte layer for the input image to remove shadows via Eq. (\ref{eq:decom}). First, we obtain a shadow mask using the shadow detector of Zhu \etal~\cite{zhu18b}. We crop the input shadow image into overlapping patches. All patches containing the shadow boundaries  are then input into the three networks. We approximate the whole image shadow parameters from the patch shadow parameters. We simply compute the image shadow parameters as a linear combination of the patch shadow parameters. Similarly, we compute the values of each pixel in the matte layer by combining the overlapping matte patches. We set the matte layer pixels in the non-shadow area to $0$ and those in the umbra area to $1$. 


%% file: Sec_6_Exp.tex
\section{Experiments}
\label{sec:exp}

\subsection{Dataset and Evaluation Metric}
We train and evaluate on the ISTD dataset \cite{Wang_2018_CVPR}. ISTD consists of image triplets: shadow image, shadow mask, and shadow-free image, captured from different scenes. The training split has 1870 image triplets from 135 scenes, whereas the testing split has 540 triplets from 45 scenes.

We notice that the testing set of the  ISTD dataset needs to be adjusted since the shadow images and the shadow-free images have inconsistent colors. This is a well known issue mentioned in the original paper \cite{Wang_2018_CVPR}. The reason is that the shadow and shadow-free image pairs were captured at different times of the day which resulted in slightly different environment lighting for each image.
For example, Fig. \ref{fig:fix_dataset} shows a shadow  and shadow-free image pair. The mean-absolute difference between these two images in the non-shadow area is 12.9. This color inconsistency appears frequently in the testing set of the ISTD dataset. On the whole testing set, the mean-absolute distance between the shadow images and shadow-free images in the non-shadow area is 6.83, as computed by Wang \etal \cite{Wang_2018_CVPR}. 

In order to mitigate this color inconsistency, we use linear regression to transform the pixel values  in the non-shadow area of each shadow-free image to map into  their counterpart values in the shadow image. We use a linear regression for each color-channel, similar to our method for relighting the shadow pixels in Sec. \ref{sec:SP-Net}. This simple transformation transfers the color tone and brightness of the shadow image to its shadow-free counterpart. The third column of Fig. \ref{fig:fix_dataset} illustrates the effect of our color-correction method. Our proposed method reduces the mean-absolute distance between the shadow-free image and the shadow image from 12.9 to 2.9. The error reduction for the whole testing set of ISTD goes from 6.83 to 2.6. We evaluate all methods on this adjusted ISTD testing set in the rest of the paper. We provide the code for our color-correction method in the project web page. 
\def\subboxsize{0.3\subFigSzab}
\begin{figure}[]
 \centering
    \includegraphics[width=0.45\textwidth]{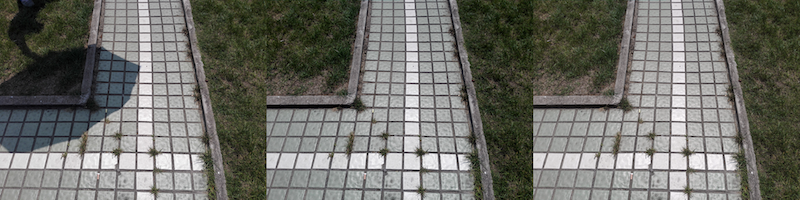}
     \makebox[\subboxsize]{Shad. Image}
    \makebox[\subboxsize]{Original GT}
    \makebox[\subboxsize]{Corrected GT}
    \caption{\textbf{An example of our color correction method.} From left to right: input shadow image, provided shadow-free ground truth image (GT) from ISTD dataset, and the GT image corrected by our method. Comparing to the input shadow image on the non-shadow area only, the mean-absolute distance of the original GT is 12.9. This value on our corrected GT becomes 2.9. 
    }
    \label{fig:fix_dataset}
\end{figure}

\subsection{Network Architectures and Implementation Details.}
We develop SP-Net by customizing a ResNeXt \cite{Xie_2017_CVPR} model. We use a U-Net architecture for M-Net and I-Net. D-Net is a simple 5-layer convolutional network. To map the outputs of the networks to their proper ranges, we use Tanh functions together with scaling and additive constants. We use stochastic gradient descent with the Adam solver \cite{Adam} to train our model. We choose a starting learning rate of $\lambda=0.0002$, which is then gradually decreased after 500 epochs. For each training iteration, the training images are resized to $320\times 320$ pixels, then randomly cropped to  $256\times256$ pixels, randomly rotated by $\alpha$ degrees where $\alpha\in[-10:10]$, and randomly flipped for data augmentation.  
All networks were trained from scratch. We experimentally set our training parameters ($\lambda_{reg}, \lambda_{sm}, \lambda_{pen},\lambda_{rec-met}, \lambda_{rec-final}$) to $(1,1, 10,1,1)$ for the fully-supervised framework and ($\lambda_{bd},\; \lambda_{mat-\alpha},\; \lambda_{sm-\alpha},\; \lambda_{adv}$) to $(0.5,\; 100,\; 10,\; 0.5)$ for the weakly-supervised framework. To choose these training parameters, we randomly split the original training set of 1330 images into a training set of 1064 images and a validation set of 266 images. Once the parameters are determined, we re-train the model with the whole training set.

We use the ISTD dataset \cite{Wang_2018_CVPR} for training. For our weakly-supervised training, each original training image of size $640\times480$ is cropped into patches of size $128\times128$ with a step size of 32. This creates 311,220 image patches from 1,330 training shadow images. This training set includes 151,327 non-shadow patches, 147,312 shadow-boundary patches, and 12,581 full-shadow patches. 

\subsection{Shadow Removal Evaluation}

We evaluate our method on the adjusted testing set of the ISTD dataset. For metric evaluation, we follow \cite{Wang_2018_CVPR} and compute the MAE in the Lab color space on the shadow area, non-shadow area, and the whole image, where all shadow removal results are re-sized into $256\times256$ to compare with the ground truth images at this size.  
Note that in contrast to other methods that only output shadow free images at that resolution, our shadow removal system works for input images of any size. 
Since our method requires shadow masks, we use the model proposed by Zhu \etal \cite{zhu18b} pre-trained on the SBU dataset \cite{Vicente-etal-ECCV16} for detecting shadows. We take the model provided by the author and fine-tune it on the ISTD dataset for 3000 epochs. This model achieves 2.2 Balanced Error Rate on the ISTD testing set. 

\subsubsection{Fully-supervised Shadow Removal Evaluation}
In Tab. \ref{tab:basic}, we compare the performance of our fully-supervised method with the fully-supervised shadow removal methods of Qu \etal\cite{Qu_2017_CVPR}, Wang \etal \cite{Wang_2018_CVPR}, Cun \etal\cite{Cun2020TowardsGS}, and Hu \etal \cite{hu_pami_2019}. All numbers are computed on the adjusted testing images so that they are directly comparable. The first row shows the numbers for the input shadow images, i.e. no shadow removal performed.

We first evaluate our shadow removal performance using only SP-Net, i.e.  we use the binary shadow mask computed by the shadow detector to form the shadow-free image from the shadow image and the relit image.  The binary shadow mask is obtained by simply thresholding the output of the shadow detector with a threshold of 0.95. As shown in  column ``\textit{SP-Net}'' (second from the right) in Fig. \ref{fig:progressive}, SP-Net correctly estimates the shadow parameters to relight the shadow area. Even with visible shadow boundaries, SP-Net alone achieves competitive shadow removal results compared to the state-of-the-art~\cite{hu_pami_2019}.

\def\subboxsize{0.24\subFigSzab}
\def\W{0.48\textwidth}
\begin{figure}[t]
 \centering
        \includegraphics[width=\W]{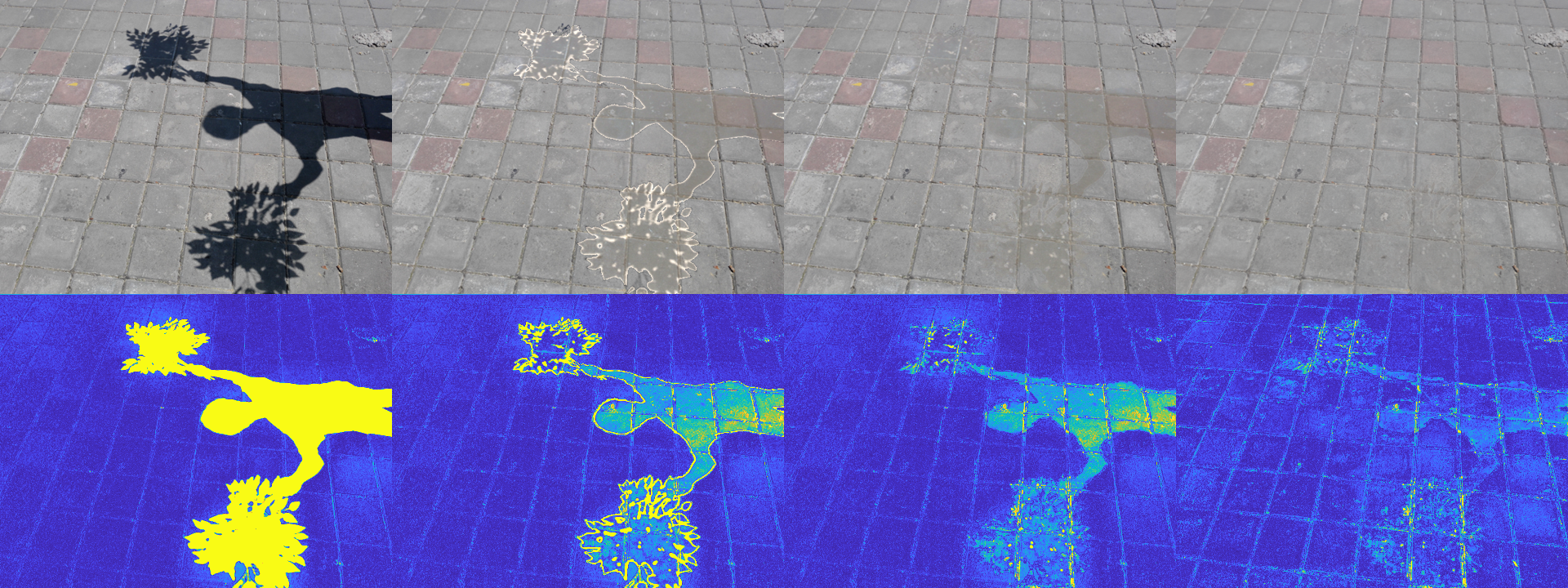}
     \makebox[\subboxsize]{Input}
    \makebox[\subboxsize]{SP-Net}
    \makebox[\subboxsize]{SP+M-Net}
    \makebox[\subboxsize]{SP+M+I-Net}
    \caption{\textbf{Shadow removal results of our model at different stages.} From left to right: input shadow images, shadow-free images obtained using  SP-Net with shadow masks from \cite{zhu18b}, shadow-free images generated by SP-Net and M-Net, and shadow-free image generated by our full system with SP-Net, M-Net, and I-Net. The bottom row visualizes the differences between each image and the ground truth shadow-free image.
    }
    \label{fig:progressive}
\end{figure}

\def\subboxsize{0.13\subFigSzab}
\def\subfig{0.95\textwidth}
\begin{figure*}[t]
 \centering

    \includegraphics[width=\subfig]{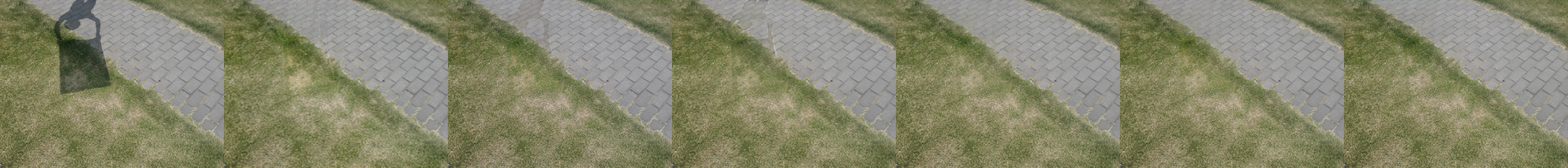}\\
    \includegraphics[width=\subfig]{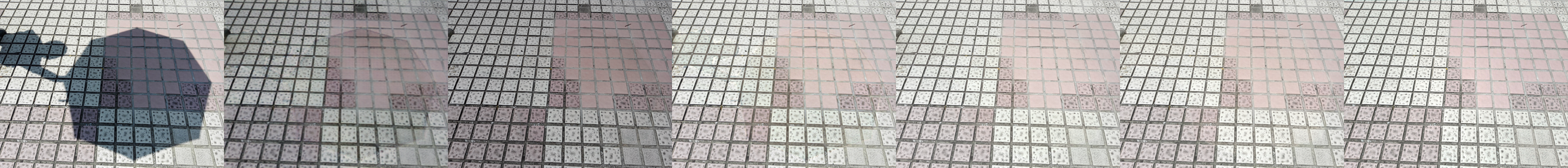}\\
    \includegraphics[width=\subfig]{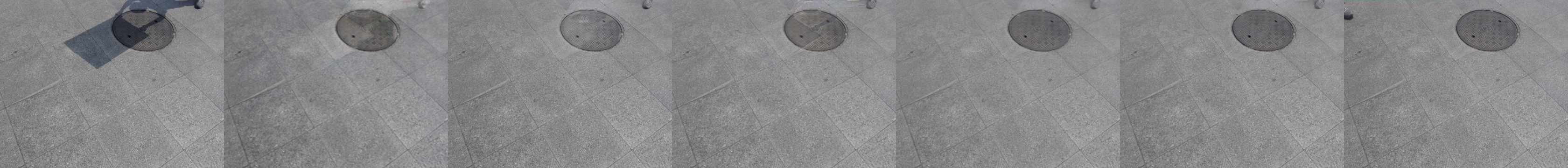}\\    
    \includegraphics[width=\subfig]{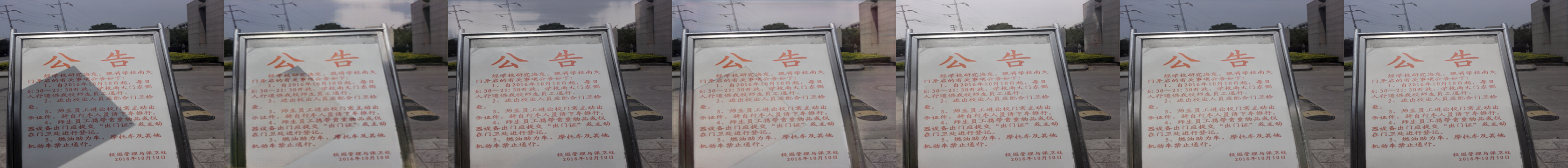}\\       
    \makebox[\subboxsize]{Input}
    \makebox[\subboxsize]{Wang \etal }
    \makebox[\subboxsize]{Cun \etal }
    \makebox[\subboxsize]{Hu \etal }
    \makebox[\subboxsize]{SP+M+I-Net}
    \makebox[\subboxsize]{SP+M+I-Net}
    \makebox[\subboxsize]{Ground }\\
    \makebox[\subboxsize]{}
    \makebox[\subboxsize]{~\cite{Wang_2018_CVPR}}
    \makebox[\subboxsize]{~\cite{Cun2020TowardsGS}}
    \makebox[\subboxsize]{~\cite{hu_pami_2019}}
    \makebox[\subboxsize]{No Shd. Mask (Ours)}
    \makebox[\subboxsize]{(Ours)}
    \makebox[\subboxsize]{Truth}
  
     \caption{\textbf{Comparison of fully-supervised shadow removal methods on the ISTD dataset.}  Qualitative comparison between our method and previous state-of-the-art methods: ST-CGAN \cite{Wang_2018_CVPR}, Cun \etal\cite{Cun2020TowardsGS}, and DSC \cite{hu_pami_2019}. ``SP+M+I-Net'' are the  shadow removal results using the parameters computed from SP-Net, the shadow matte computed from M-Net, and the residual layer computed from I-Net. 
    }
 
    \label{fig:main}
\end{figure*}

\input{main_table}

We then evaluate the shadow removal results using both SP-Net and M-Net, denoted as 
``\textit{SP+M-Net}'' in Tab. \ref{tab:basic}. As shown in Fig. \ref{fig:progressive}, the results of SP+M-Net do not contain  boundary artifacts. Moreover, M-Net also tends to correct some pixels that are over-lit by SP-Net. As M-Net is trained to blend the relit and shadow images  to create the shadow-free image, it learns to output a smaller weight for a pixel that is over-lit by SP-Net. Using the matte layer of M-Net further reduces the MAE on the shadow area by 30\%, from 8.4 to 6.5, outperforming the method of Hu \etal\cite{Hu_2018_CVPR} by 14\%. 
Our inpainting network, I-Net, brings another 8\% MAE error reduction, helping our full system to achieve a 6.0 MAE on the shadow areas of the ISTD dataset. Lastly, our method achieves a 5.3 MAE on the shadow areas when using the ground-truth shadow masks.

\def\subboxsize{0.27\subFigSzab}
\begin{figure}[]
 \centering

    \includegraphics[width=0.45\textwidth]{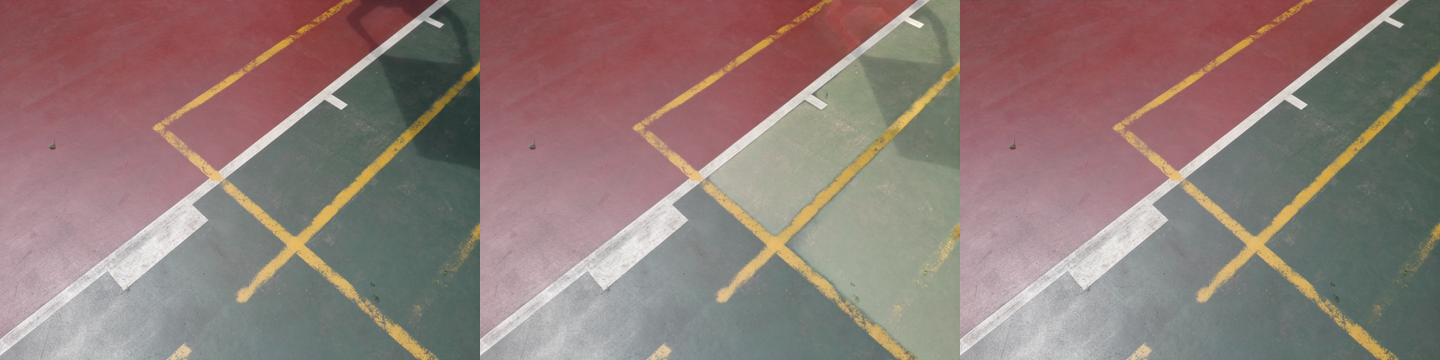}
    \includegraphics[width=0.45\textwidth]{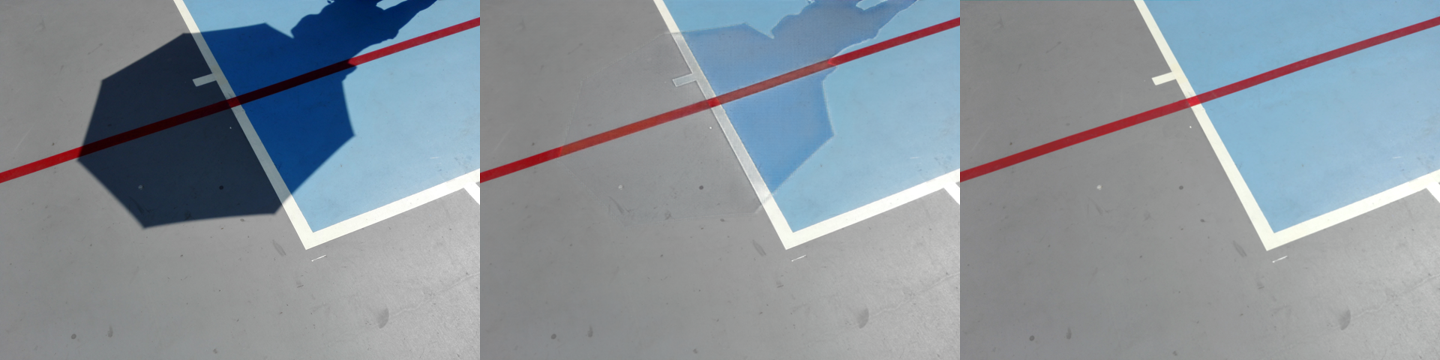}
     \makebox[\subboxsize]{Input}
    \makebox[\subboxsize]{Ours}
        \makebox[\subboxsize]{GT}
 
    \caption{\textbf{Failure cases of our method.} 
     The first row shows a case where our method fails to accurately localize the shadow area while in the second row, the color of the lit area is incorrect. 
    }
    \label{fig:fail}
\end{figure}

\def\subboxsize{0.15\textwidth}
\def\subfig{0.45\textwidth}
\def\W{0.48\textwidth}
\def\H{0.13\textwidth}
\begin{figure*}[]
 \centering

    \includegraphics[width=\W, height=\H]{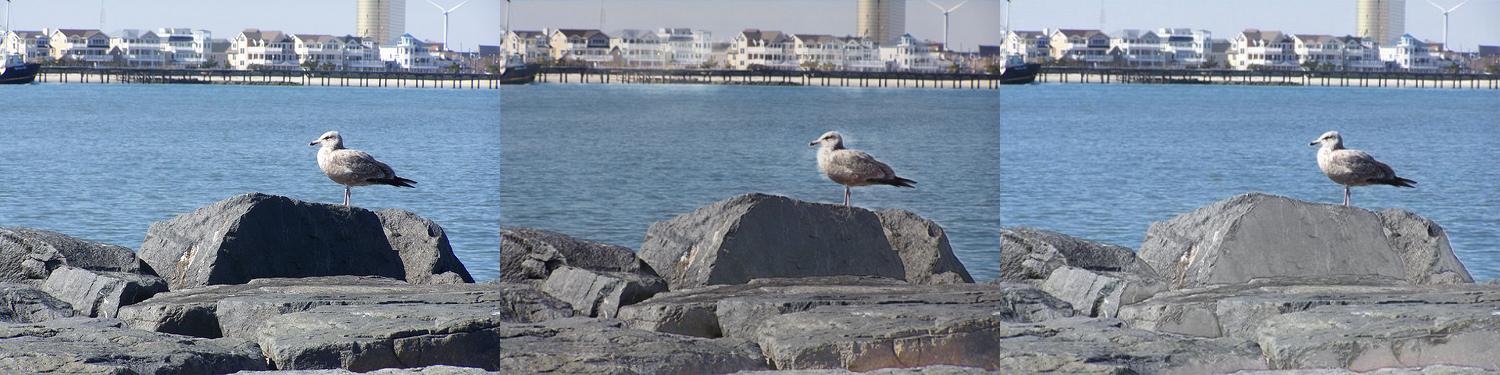}
    \includegraphics[width=\W, height=\H]{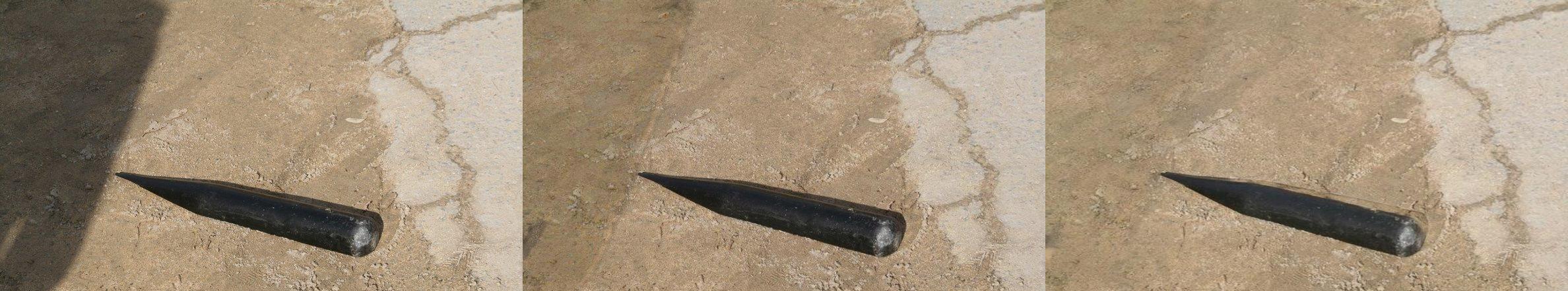}
    \includegraphics[width=\W, height=\H]{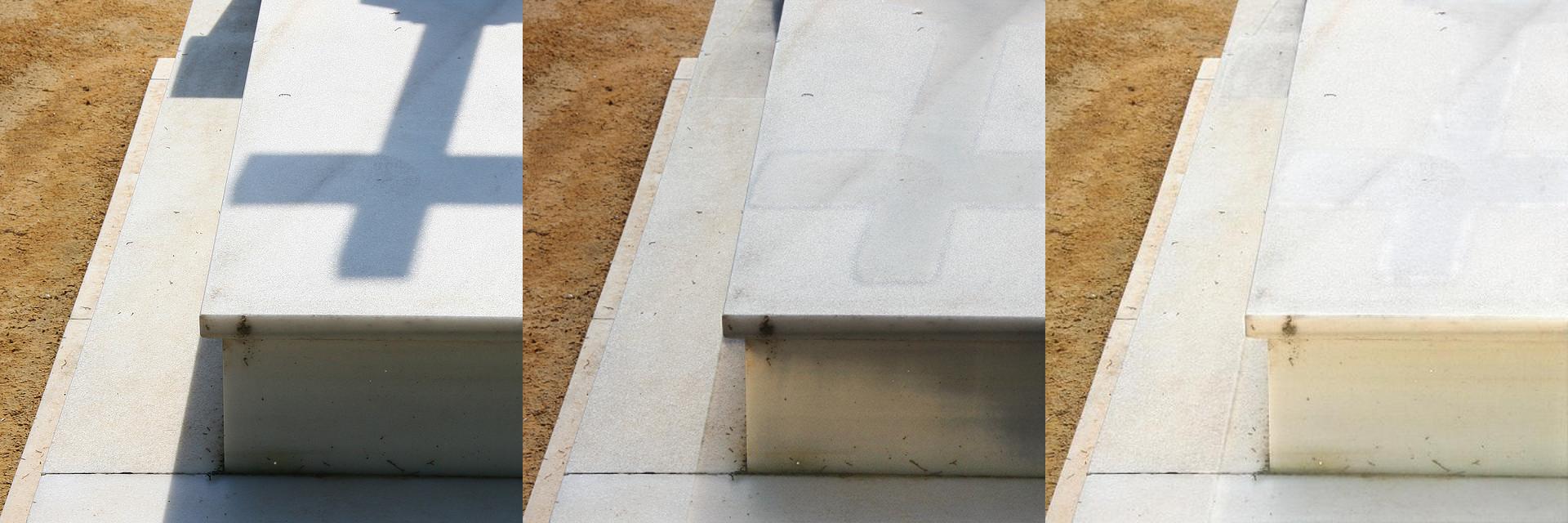}
    \includegraphics[width=\W, height=\H]{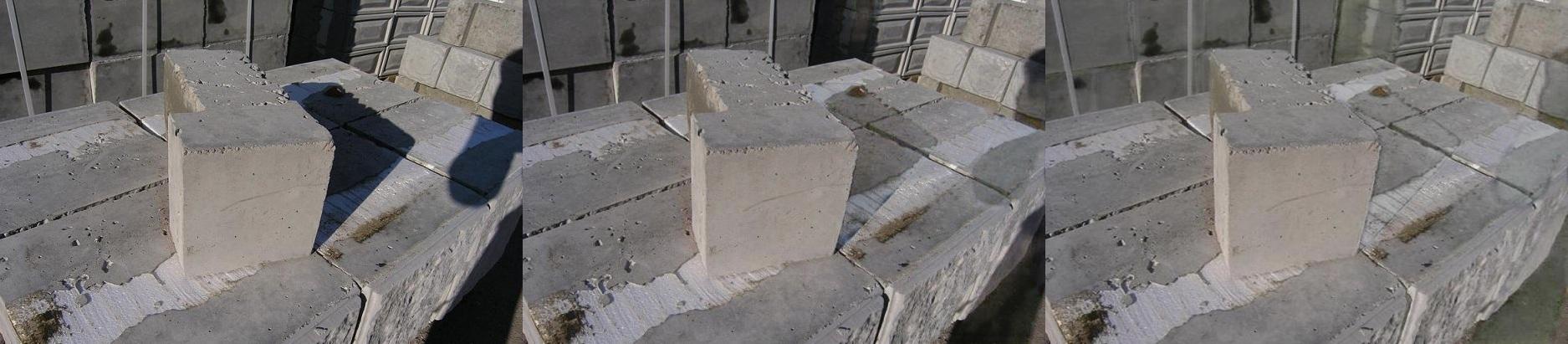}
    \includegraphics[width=\W, height=\H]{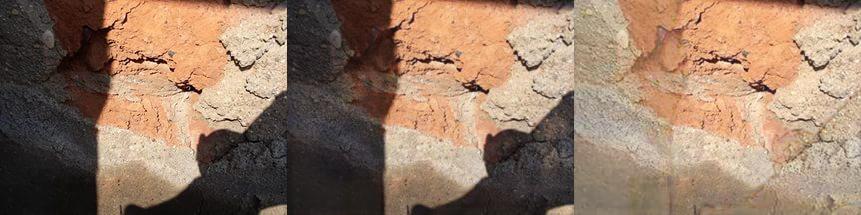}
    \includegraphics[width=\W, height=\H]{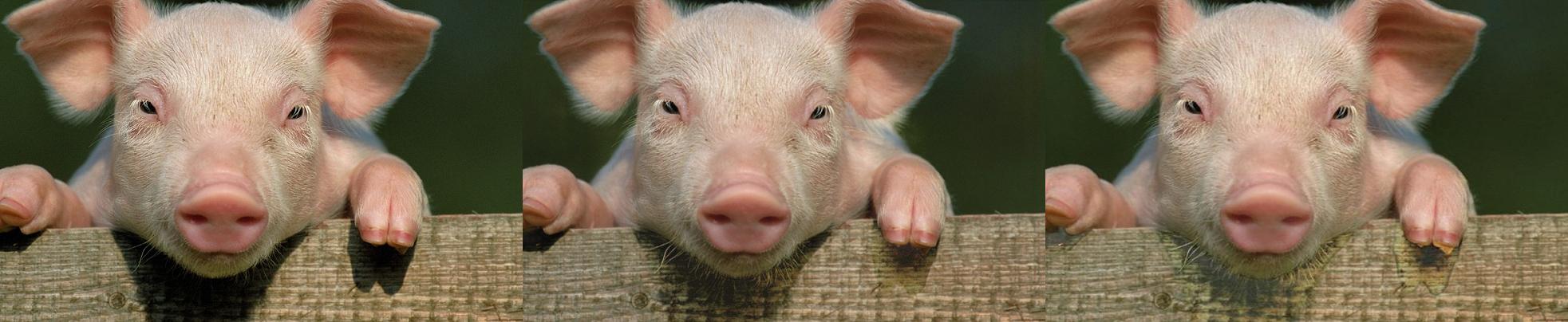}
    \includegraphics[width=\W, height=\H]{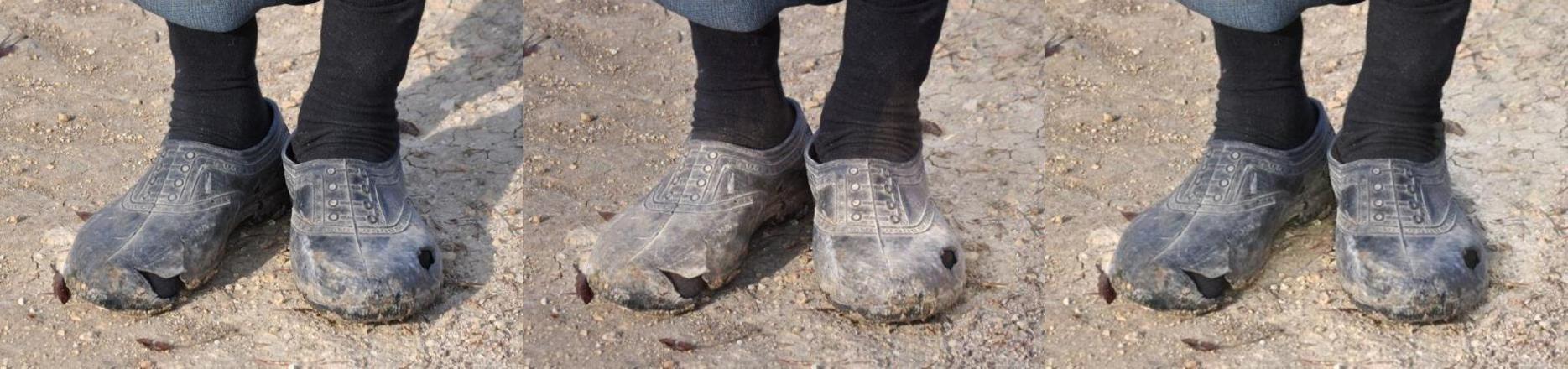}
    \includegraphics[width=\W, height=\H]{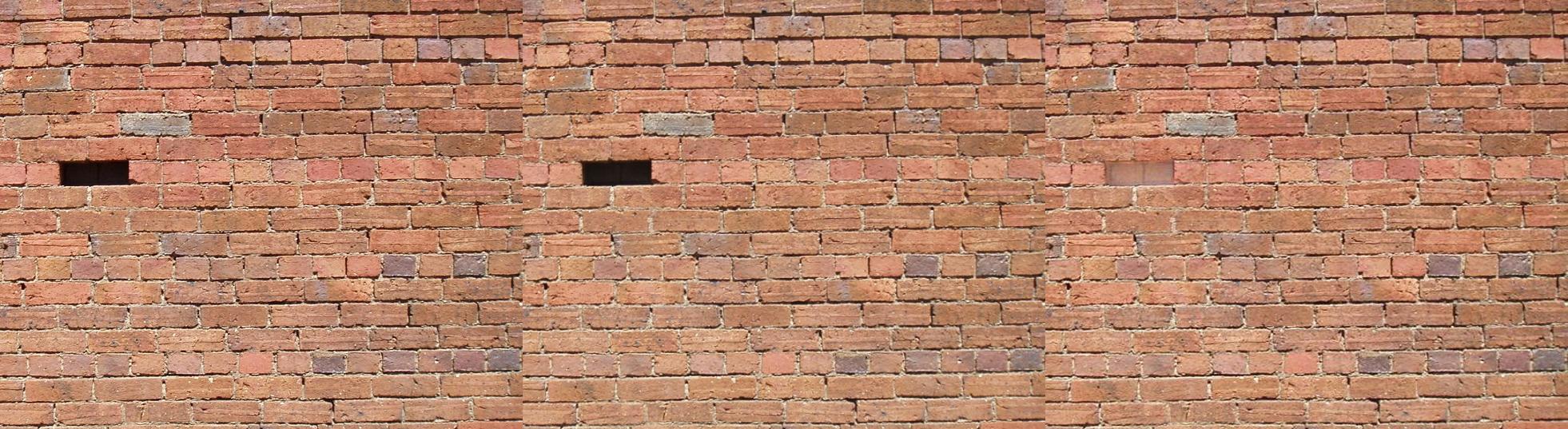}

    \makebox[\subboxsize]{Input}
    \makebox[\subboxsize]{Cun \etal }
    \makebox[\subboxsize]{SP+M+I-Net}
    \makebox[\subboxsize]{Input}
    \makebox[\subboxsize]{Cun \etal }
    \makebox[\subboxsize]{SP+M+I-Net}
  
     \caption{\textbf{Comparison between our shadow removal method and the method of Cun \etal \cite{Cun2020TowardsGS} on the SBU dataset\cite{Vicente-etal-ECCV16} 
    }}
 
    \label{fig:SBU_1}
\end{figure*}

Fig. \ref{fig:main} visualizes the shadow removal results of our method in comparison with previous state-of-the-art methods including ST-CGAN \cite{Wang_2018_CVPR}, the method of Cun \etal\cite{Cun2020TowardsGS}, and DSC \cite{hu_pami_2019}. Note that all methods are based on deep-learning and are trained with paired data. Overall, our method generates better results without image artifacts. The methods of Wang \etal \cite{Wang_2018_CVPR} and Cun \etal \cite{Cun2020TowardsGS} tend to darken the overall brightness of the images. There are visual artifacts in all other methods, especially in areas with dark materials. There are visible boundary artifacts in the outputs of all methods even though they are milder in our case. 

Our method has some failure cases,  visualized in Fig. \ref{fig:fail}. The first row shows a case where our method fails to accurately localize the shadow area due to the inaccurate predicted shadow mask while in the second row, the color of the lit area is incorrect due to the inaccurate estimated shadow parameters.

Additionally, we trained and evaluated an alternative design that does not input the shadow masks into SP-Net, M-Net, and I-Net. Hence, SP-Net, M-Net, and I-Net need to learn to localize the shadow areas implicitly. As can be seen in the three bottom rows of  Tab. \ref{tab:basic}, this design performs slightly worse than our main setting but still outperforms previous state-of-the-art methods by a large margin. Some examples are visualized in Fig. \ref{fig:main}.

Fig.\ref{fig:SBU_1} visualizes shadow removal results of our method in comparison to the shadow removal method of Cun \etal\cite{Cun2020TowardsGS} on the SBU dataset\cite{Vicente-etal-ECCV16}, which includes images with complex shadows on various scenes. Both our method, SP+M+I-Net, and the method of Cun \etal~  are trained on the ISTD dataset that only includes images with simple scenes and shadows. Thus, this comparison evaluates how the methods generalize to unseen and complex shadows that are present in natural images.
As can be seen, the method of Cun \etal~ only attenuates the shadows in most cases and tends to modify the overall brightness of the images.  Our method removes shadows in a more consistent manner albeit with the visible boundary artifacts in some cases.

\subsubsection{Weakly-supervised Shadow Removal Evaluation}
We compare our weakly-supervised approach with earlier prior-based and weakly-supervised shadow removal methods in Tab. \ref{tab:weakly}. The methods of Yang \etal\cite{Yang12} and  Gong~\etal\cite{Gong16} are prior-based while the method of Hu \etal \cite{hu_iccv2019mask} is a weakly-supervised deep-learning method based on CycleGAN \cite{CycleGAN2017}. While our method does not require any shadow-free images to train, Mask-ShadowGAN requires unpaired shadow and shadow-free images. Our method outperforms Mask-ShadowGAN by 22\%, reducing the MAE in the shadow area from 12.4 to 9.7 while also achieving lower MAE on the non-shadow area. Note that we even outperform some fully-supervised methods whose numbers are reported in Tab.\ref{tab:basic}. Our method achieves slightly better results when using the ground-truth shadow masks, shown in the last row of Tab. \ref{tab:weakly}.

Fig. \ref{fig:weakly} visualizes the shadow removal results of our weakly-supervised method, denoted as "wSP+M-Net", in comparison with Mask-ShadowGAN. Mask-ShadowGAN outputs images with visible undesired artifacts even for relatively simple cases. The shadowed pixels are usually over-lit and the textures beneath the shadows are not reconstructed properly. Our method handles most cases relatively well. Compared to our fully-supervised method, our weakly-supervised method produces more boundary artifacts.

\def\subboxsize{0.22\subFigSzab}
\def\subfig{0.45\textwidth}
\begin{figure}[]
 \centering

    \includegraphics[width=\subfig]{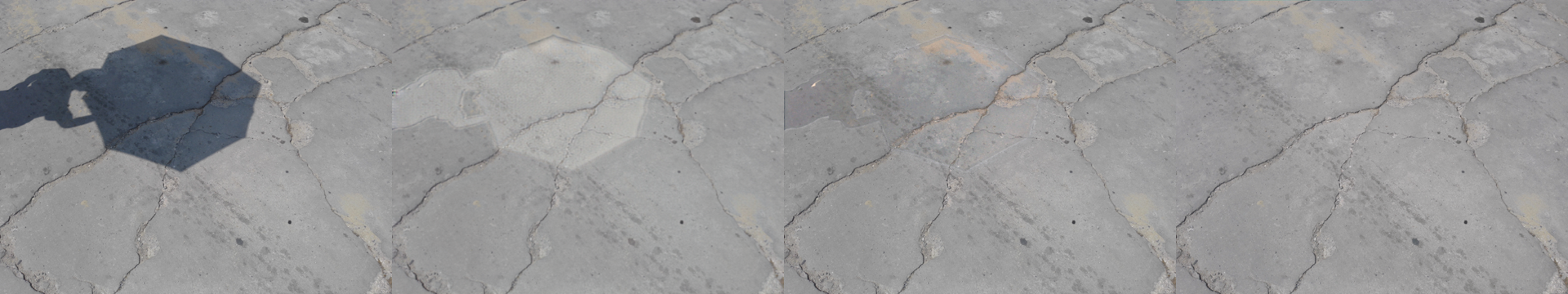}\\
    \includegraphics[width=\subfig]{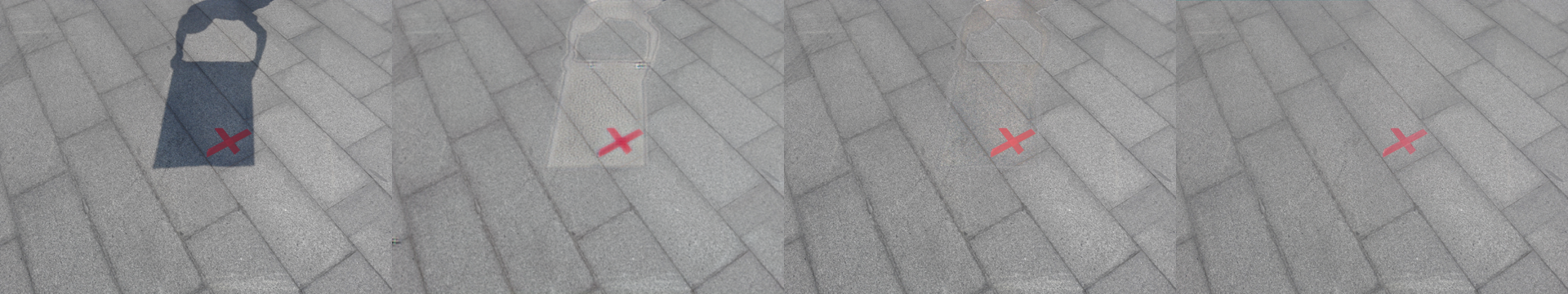}\\
    \includegraphics[width=\subfig]{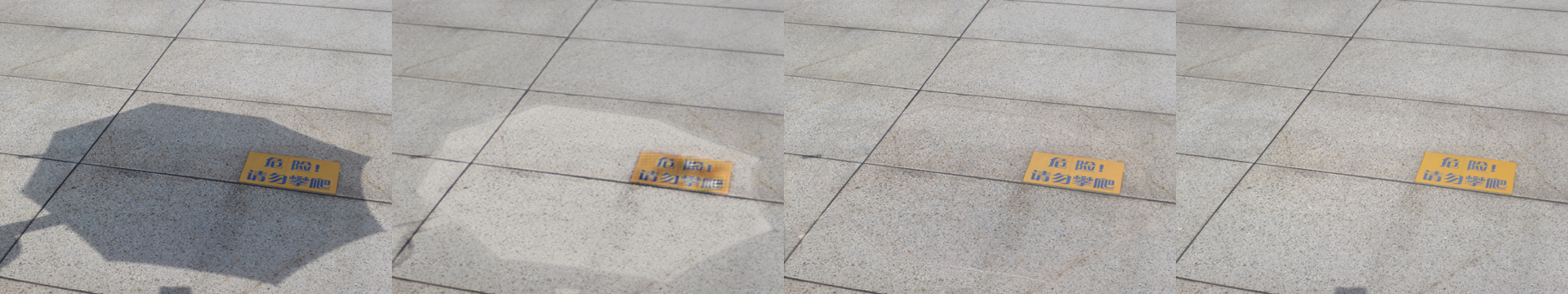}\\
    \includegraphics[width=\subfig]{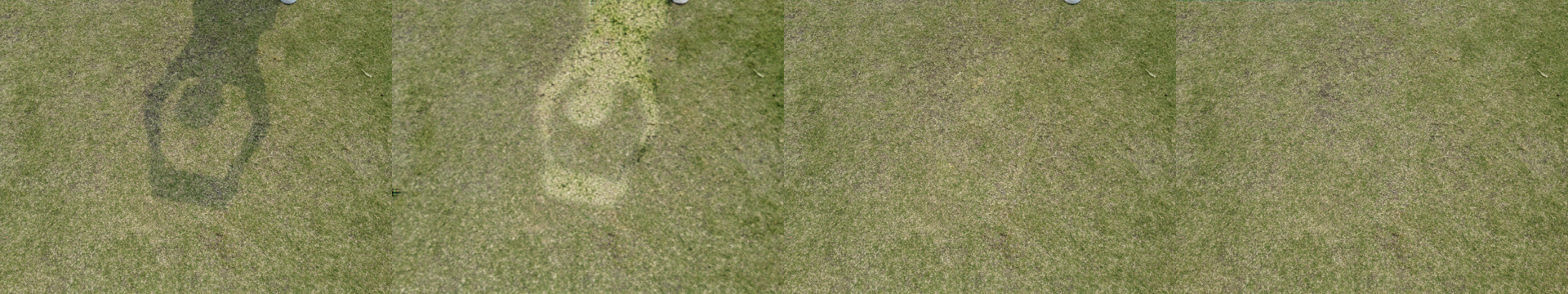}\\
    \makebox[\subboxsize]{Input}
    \makebox[\subboxsize]{MaskShad.}
    \makebox[\subboxsize]{wSP+M-Net}
    \makebox[\subboxsize]{Ground }\\
    \makebox[\subboxsize]{}
    \makebox[\subboxsize]{-GAN~\cite{hu_iccv2019mask}}
    \makebox[\subboxsize]{(Ours)}
    \makebox[\subboxsize]{Truth}
  
     \caption{\textbf{Qualitative comparison between our method and the previous state-of-the-art weakly-supervised method Mask-ShadowGAN \cite{hu_iccv2019mask}.}   ``wSP+M-Net'' are the  shadow removal results of our weakly-supervised method trained using only shadow images and shadow masks. 
    }
 
    \label{fig:weakly}
\end{figure}

\subsection{Ablation Studies}
\label{sec:ablation}

\setlength{\tabcolsep}{4pt}
\begin{table}[h!]
\begin{center}
\caption{\textbf{Ablation Studies.} We train our networks with different settings and report the shadow removal performances on the ISTD dataset \cite{Wang_2018_CVPR}. The metric is MAE (the lower, the better). }
\label{table:abl}
\begin{tabular}{lccc}
\hline\noalign{\smallskip}
Methods       & Shadow& Non-Shadow& All  \\ 
\noalign{\smallskip}
\midrule
\noalign{\smallskip}
Input Image  & 40.2  & 2.6 & 8.5\\ 
\midrule
\midrule
\multicolumn{4}{c}{Fully-Supervised Framework}\\
\midrule
SP+M-Net (Baseline)   &7.9  &3.1&3.9\\
SP+M-Net + L.S.S  &7.1  &3.0 &3.7\\ 
SP+M-Net + L.S.S + $\mL_{pen}$    & 6.7  &3.1 &3.6\\
SP+M+I-Net + L.S.S + $\mL_{pen}$   & 6.2  &3.1 &3.6\\
\midrule
SP+M+I-Net + L.S.S  + $\mL_{pen}$ + $\mL_{sm-\alpha}$   & 6.0  &3.1 &3.6\\
\midrule
\midrule
\multicolumn{4}{c}{Weakly-Supervised Framework}\\
\midrule
wSP+M-Net w/o L.S.S  &47.5  &2.9&9.9\\
wSP+M-Net w/o $\mL_{bd}$  &41.7  &3.9 &9.8\\ 
wSP+M-Net w/o $\mL_{mat-\alpha}$  & 38.7 &3.1 &9.0\\
wSP+M-Net w/o $\mL_{sm-\alpha}$  & 10.2  &2.8 &4.0\\
wSP+M-Net w/o $\mL_{GAN}$ & 26.9  &2.9 &6.8\\
\midrule
wSP+M-Net  & 9.7  &3.0 &4.0\\

\midrule
\end{tabular}
\end{center}
\end{table}

We conduct ablation studies to better understand the effects of each proposed component in our framework in both fully-supervised and weakly-supervised settings. The results are summarized in Tab.\ref{table:abl}.

For the fully-supervised setting, we start from a baseline model \cite{Le-etal-ICCV19} that is trained without any proposed components, i.e., limiting the search space of SP-Net and M-Net (denoted as ``L.S.S''), the penumbra reconstruction loss ($L_{pen}$), the inpainting network I-Net, and the smoothness loss ($L_{sm-\alpha}$). As can be seen in the second row, simply constraining the values of the shadow parameters and shadow mattes to their proper ranges significantly improves the shadow removal performance by a 10\% error reduction in terms of MAE on the shadow areas, from 7.9 to 7.1. The penumbra reconstruction loss guides M-Net to focus more on the penumbra areas of shadows. Incorporating this loss into our framework improves the shadow removal performance by 6\%. 
The addition of the inpainting network, I-Net, brings another 7\% error reduction. The last component is a smoothness loss that regularizes the matte layer, which reduces the MAE to 6.0.

For the weakly-supervised method, we show that all proposed features and loss functions are crucial in training our framework. Starting from the original model, we train new  models removing the proposed components one at a time. 
 The first row shows the results of our model when we set the search space of the scaling factor $w$ to $[-10,10]$ and the search space of the additive constant $b$ to $[-255,255]$. In this case, the model collapses and consistently outputs uniformly dark images. Similarly, the model collapses when we omit the boundary loss $\mL_{bd}$. We observe that this loss is essential in stabilizing the training as it prevents the Param-Net from outputting consistently high values. 
\par The matting loss $\mL_{mat-\alpha}$ and the $\mL_{GAN}$ loss are critical for learning  proper shadow removal. We observe that without the matting loss $\mL_{mat-\alpha}$, the model behaves similarly to an image inpainting model where it tends to modify all parts of the images to fool the discriminator.
Last, dropping the smoothness loss $\mL_{sm}$ only results in a  slight drop in shadow removal performance, from 9.7 to 10.2  MAE on the shadow areas. However, we observe more visible boundary artifacts on the output images without this loss. 

\subsection{SBU-Timelapse Dataset}
\label{sec:dataset}

Paired shadow data for complex scenes and complex shadows are difficult to collect. Current shadow removal datasets only contain images with simple scenes, simple shadows, and without the oclluders in the images due to the data acquisition scheme \cite{Qu_2017_CVPR, Wang_2018_CVPR} for paired shadow and non-shadow images.

We propose a method to obtain paired shadow data based on time-lapse videos. We collect SBU-Timelapse, a video dataset of 50 videos, each contains a static scene without visible moving objects. We cropped those videos to obtain clips with the only dominant motions caused by the shadows (either by  direct light motion or motion of the unseen occluders). Instead of collecting paired shadow and shadow-free images, we aim to collect paired shadow and non-shadow pixels. As the shadows travel across the scene, we can obtain paired shadow and non-shadow values for pixels that go in or out of the shadows. As this method does not require interacting with the occluders or capturing the shadow-free images, it allows collecting images with complex shadows, complex backgrounds, or self-cast shadows, as can be seen from Fig. \ref{fig:video_exp}. 

Inspired by \cite{Chuang2003}, we propose a ``max-min'' technique to obtain a single pseudo shadow-free frame for each video: since the camera is static and there is no visible moving object in the frames, the changes in the video are caused by the moving shadows. We first obtain two images $V_{max}$ and $V_{min}$ by taking the maximum and minimum intensity values at each pixel location across the whole video. $V_{max}$ is then the image that contains the shadow-free values of pixels if they ever go out of the shadows. Similarly, their shadowed values, if they ever go into the shadows, are captured in $V_{min}$. Fig. \ref{fig:video_anno} shows these two images for an example video. From these two images, we can trivially obtain a mask, namely moving-shadow $\mM$, marking the pixels appearing in both the shadow and non-shadow areas in the video: 

\begin{align}
	\mM_{i} = \left\{ 
	\begin{array}{ll}
		1 & \hspace{3ex}\textrm{if } V_{max,i}  > V_{min,i}  + \epsilon \\
		0 & \hspace{3ex}\textrm{otherwise},
	\end{array}
	\right. 
\end{align}
where we set a threshold of $\epsilon=80$.  This method allows us to obtain pairs of shadow and non-shadow pixel values in the moving-shadow mask, $\mM$, for free. Last, we manually cleaned the obtained moving-shadow mask to eliminate the areas of changes that are not associated to shadows, e.g., changes due to subtle motions or color shifts in the videos. All the original video clips, extracted frames, adjusted moving-shadow masks, and evaluation code are available as the SBU-Timelapse dataset.

\def\subfig{0.18\textwidth}
\def\subfigH{0.12\textwidth}
\def\subboxsize{0.18\textwidth}
\begin{figure*}[t]
 \centering
\includegraphics[width=\subfig,height=\subfigH]{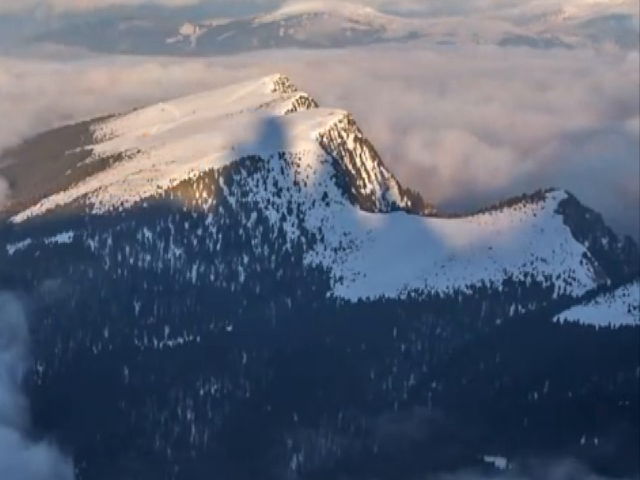}
\includegraphics[width=\subfig,height=\subfigH]{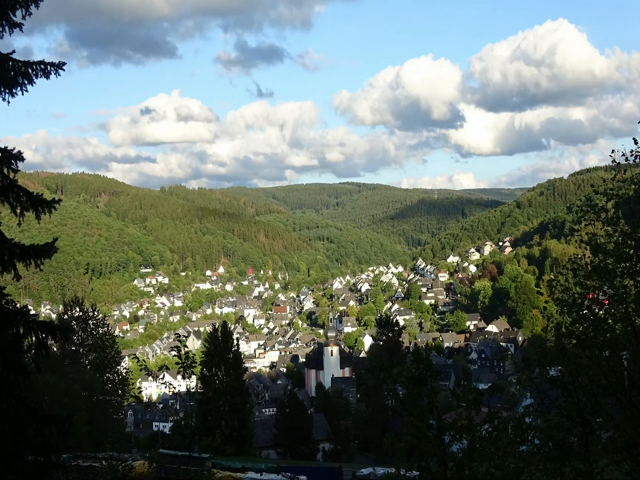}
\includegraphics[width=\subfig,height=\subfigH]{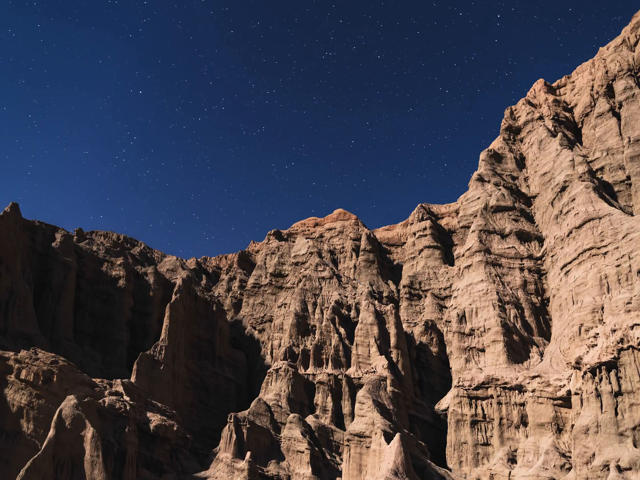}
\includegraphics[width=\subfig,height=\subfigH]{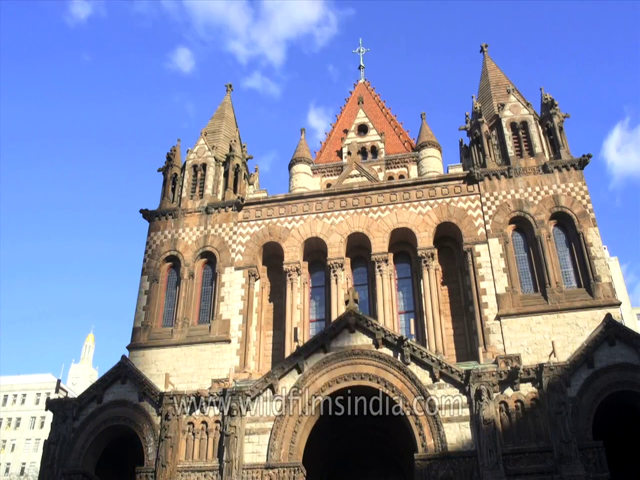}
\includegraphics[width=\subfig,height=\subfigH]{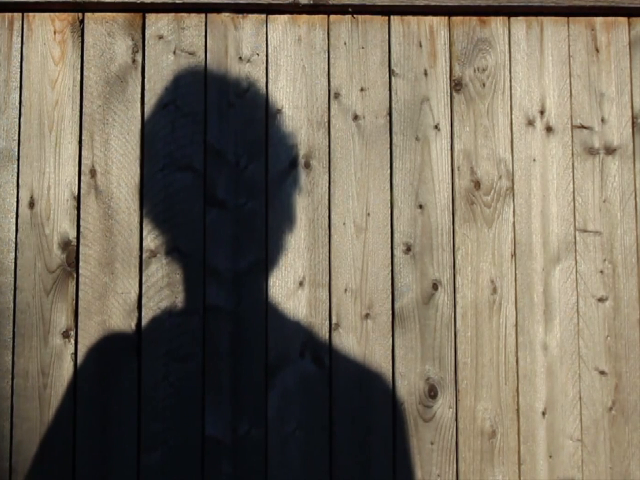}\\
\includegraphics[width=\subfig,height=\subfigH]{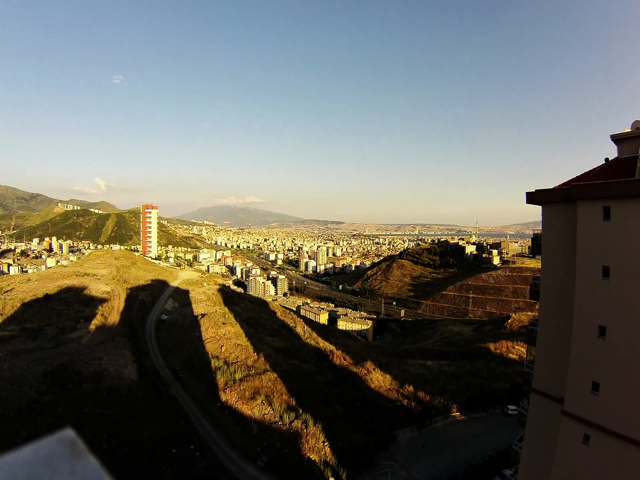}
\includegraphics[width=\subfig,height=\subfigH]{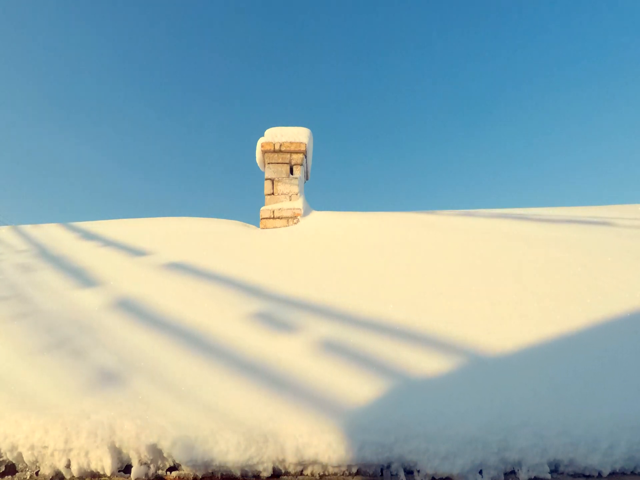}
\includegraphics[width=\subfig,height=\subfigH]{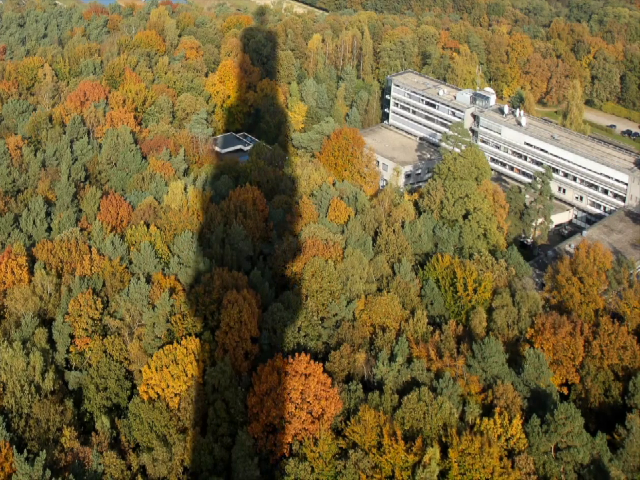}
\includegraphics[width=\subfig,height=\subfigH]{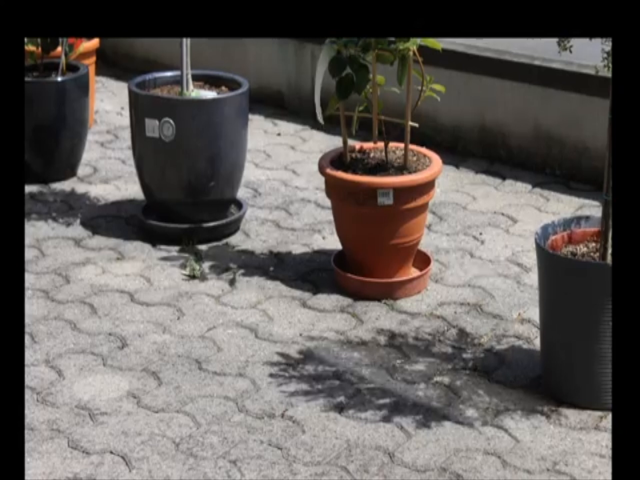}
\includegraphics[width=\subfig,height=\subfigH]{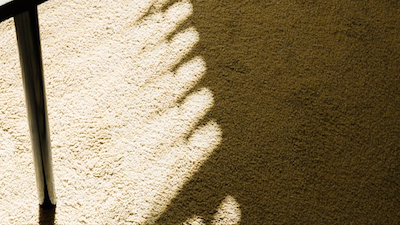}\\

     \caption{{\bf Examples of the SBU-Timelapse dataset.} We collected a time-lapse video dataset where both the scene and the visible objects remain static. The dataset includes videos containing shadows cast by close-up occluders, far distance occluders, videos with simple-to-complex shadows, and shadows on various types of backgrounds and materials.
    }
    \label{fig:video_exp}
\end{figure*}

\def\subfig{0.18\textwidth}
\def\subfigH{0.12\textwidth}
\def\subboxsize{0.18\textwidth}
\begin{figure*}[t]
 
 \centering

\includegraphics[width=\subfig,height=\subfigH]{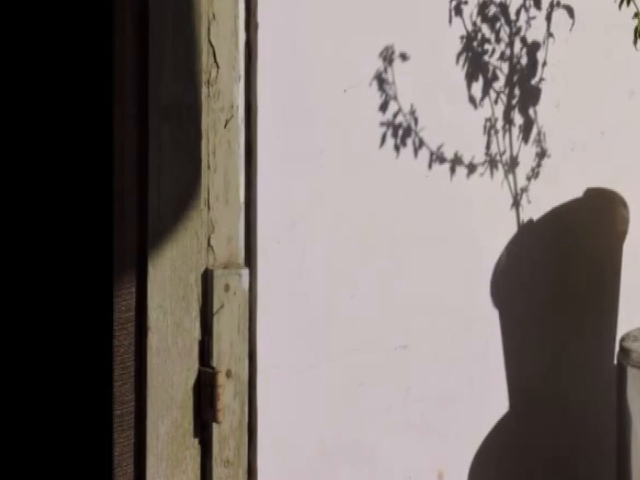}
\includegraphics[width=\subfig,height=\subfigH]{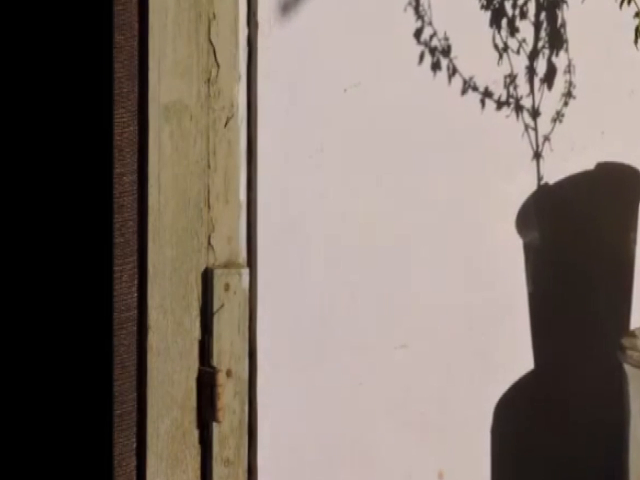}
\includegraphics[width=\subfig,height=\subfigH]{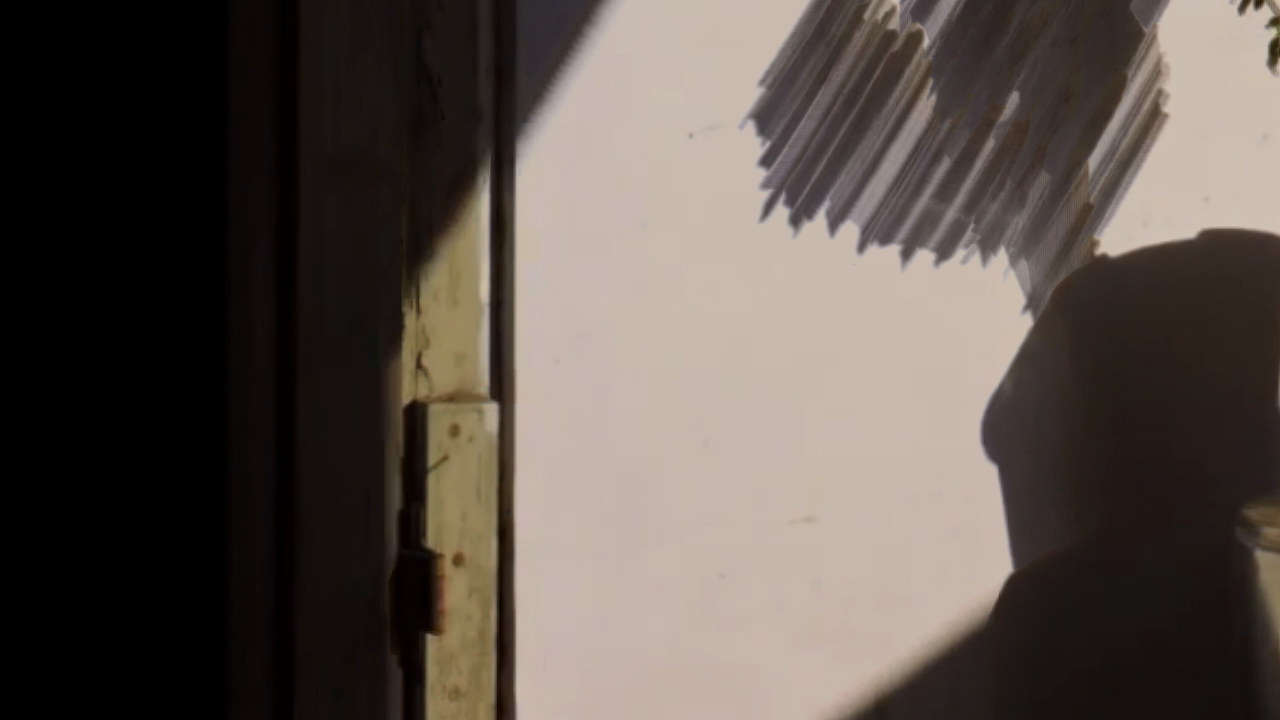}
\includegraphics[width=\subfig,height=\subfigH]{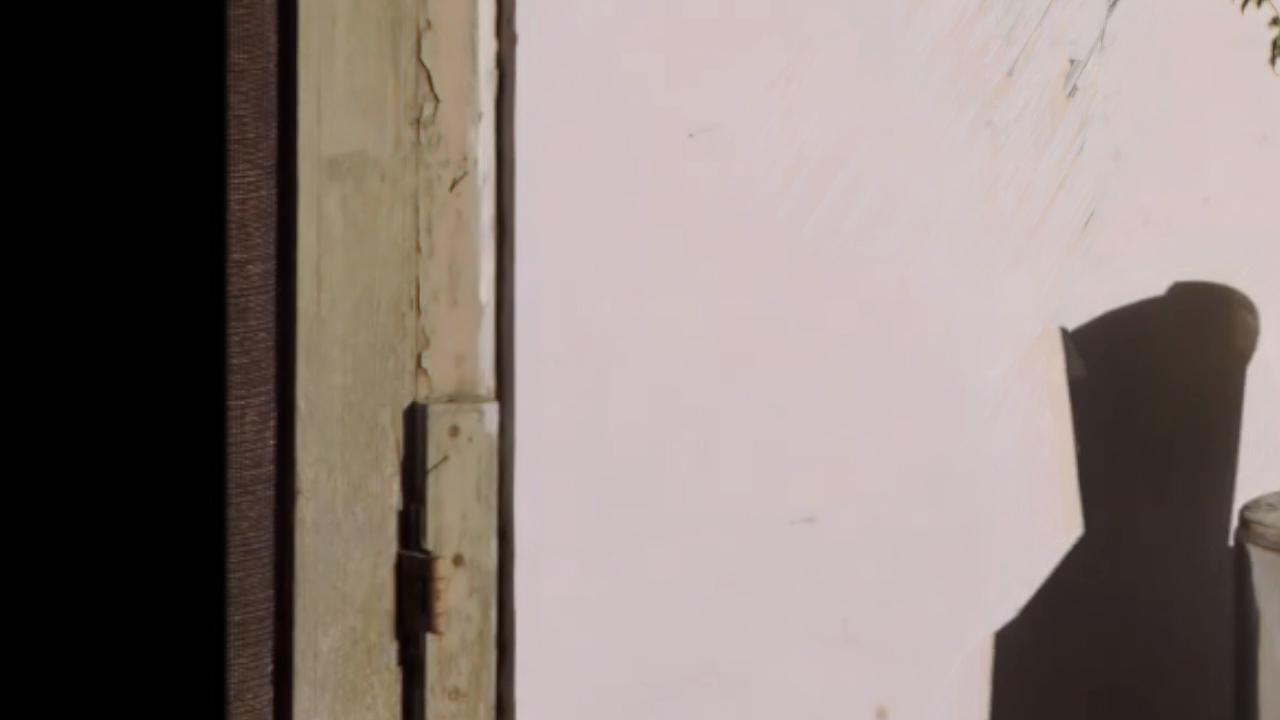}
\includegraphics[width=\subfig,height=\subfigH]{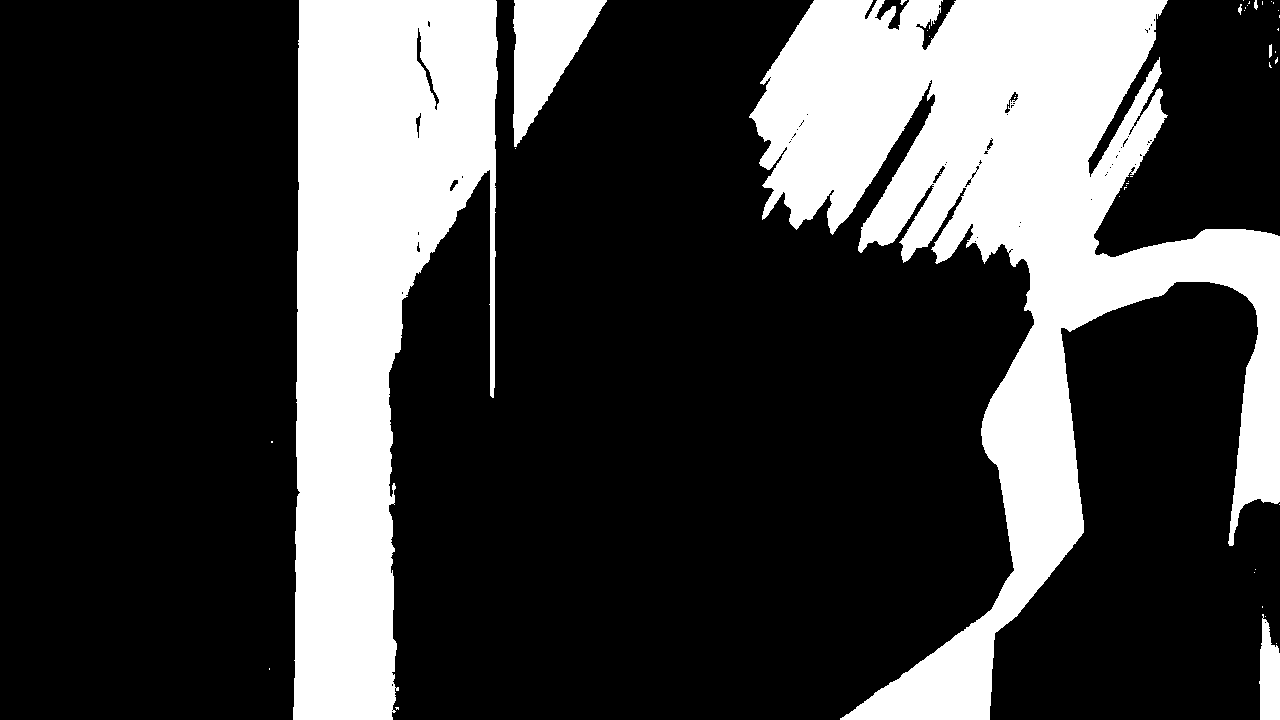}

     \makebox[\subboxsize]{Frame 0}
    \makebox[\subboxsize]{Frame 100}
    \makebox[\subboxsize]{$V_{min}$ }
    \makebox[\subboxsize]{$V_{max}$ }
    \makebox[\subboxsize]{Moving Shadow}\\
     \caption{{\bf Our method to obtain pseudo shadow-free frames and moving-shadow masks for  shadow removal evaluation.} We first obtain $V_{max}$ and $V_{min}$ by taking the maximum and minimum values at each pixel location across the whole video. We use $V_{max}$ as a single pseudo shadow-free ground truth frame for the whole video. The moving-shadow masks can be easily obtained by comparing the $V_{max}$ and $V_{min}$ images. We measure the shadow removal performance only on the moving shadow mask areas to exclude the pixels that are constantly in shadow for the whole video.  
    }
    \label{fig:video_anno}
\end{figure*}

To measure shadow removal performance, we input the frames of these videos into the shadow removal algorithm and measure the MAE in the Lab color space between the output frame and the image $V_{max}$ on the moving-shadow area $\mM$. We compute MAE on each video and take their average to measure the shadow removal performance on the whole dataset. 

Tab. \ref{table:video} summarizes the performance of our methods compared to Mask-ShadowGAN\cite{hu_iccv2019mask} and the method of Cun \etal \cite{Cun2020TowardsGS} on these videos. Our methods generalize better than other methods on the time-lapse video dataset.
As can be seen, the shadow removal method of Cun \etal\cite{Cun2020TowardsGS} does not generalize well to this test where it only reduces the MAE on the moving-shadow areas by 8\%, compared to the input frames. Our weakly-supervised method outperforms the method of Cun \etal\cite{Cun2020TowardsGS} and Mask-ShadowGAN, reducing the MAE by 25\% and 11\% respectively. Our method trained with fully-supervised data achieves the lowest error rate among all methods. However, it does not show significant improvement over its weakly-supervised counterpart and Mask-ShadowGAN. These results suggest that
  data-driven methods trained on a dataset with limited diversity fail to generalize to other scenes with complex shadows. Note that while the ISTD dataset mostly contains images with only outdoor scenes with a single shadow instance per image, shadows presented in our video shadow dataset are significantly more diverse, as shown in Fig.\ref{fig:video_exp}. 
  
  Some shadow removal examples for a single frame per video are shown in Fig.\ref{fig:main_video}. There are challenging cases with complex shadows where all methods fail to remove shadows properly. The method of Cun \etal~and Mask-ShadowGAN tend to introduce artifacts and are not able to localize the shadow areas correctly. Our method certainly has limitations: it inherits the error from the shadow detector (3rd row) and cannot deal with multiple shadow instances in the same image (4th row).

\setlength{\tabcolsep}{4pt}
\begin{table}[t]
\begin{center}
\caption{\textbf{Shadow removal results on our proposed SBU-Timelapse dataset}. The metric is MAE (the lower, the better), compared to the pseudo shadow-free frame on the moving shadow mask.} 
\label{table:video}
\begin{tabular}{lcc}
\hline\noalign{\smallskip}
Methods & Training Data (ISTD Dataset) &MAE \\
\noalign{\smallskip}
\midrule
\noalign{\smallskip}
Input Frames &- & 33.8 \\
\midrule
Hu \etal \cite{hu_iccv2019mask} & Unpaired Shadow/Shadow-Free   & 27.3 \\
Cun \etal \cite{Cun2020TowardsGS} & Paired Shadow/Shadow-Free  & 31.2 \\
\midrule
wSP+M-Net (Weakly-trained) & Shadow Images & 23.4 \\
SP+M+I-Net (Fully-trained) & Paired Shadow/Shadow-Free  & \textbf{20.1 }\\
\midrule
\end{tabular}
\end{center}
\end{table}

\def\subboxsize{0.13\subFigSzab}
\def\subfig{0.95\textwidth}
\begin{figure*}[]
 \centering

    \includegraphics[width=\subfig]{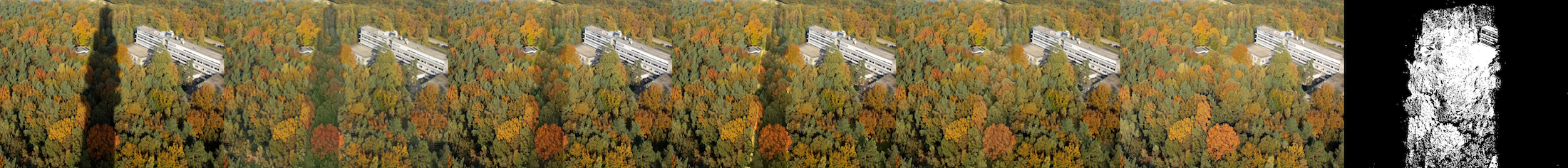}\\
    \includegraphics[width=\subfig]{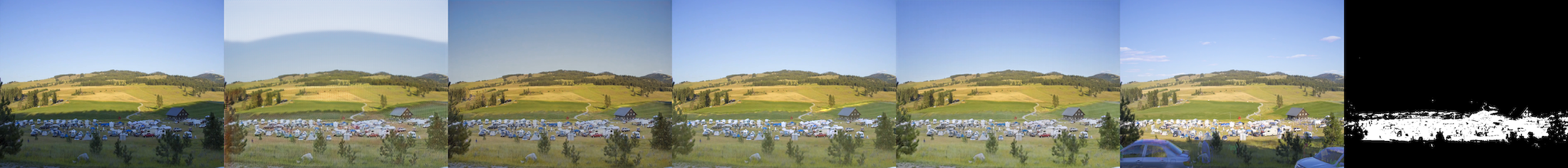}\\
        \includegraphics[width=\subfig]{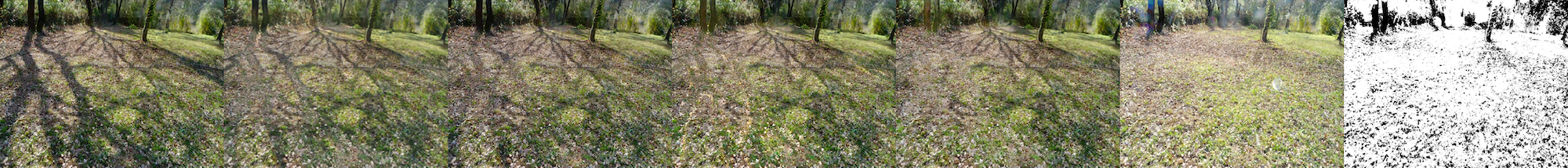}\\
    \includegraphics[width=\subfig]{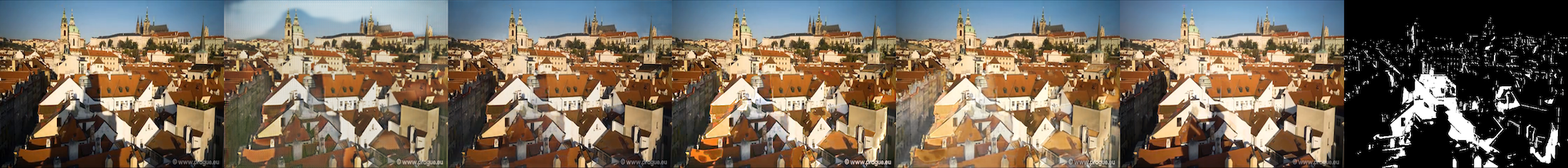}\\

     \includegraphics[width=\subfig]{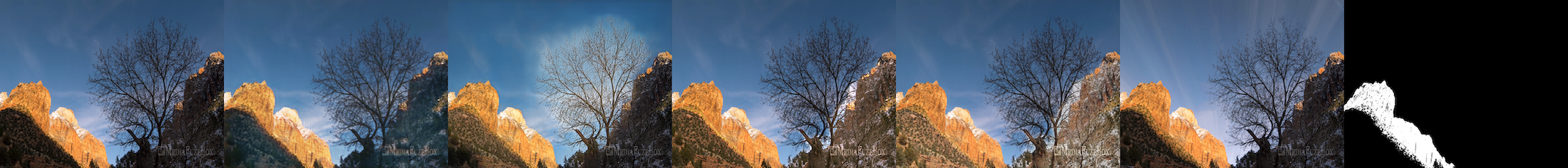}\\
    \makebox[\subboxsize]{Input}
    \makebox[\subboxsize]{Hu \etal }
    \makebox[\subboxsize]{Cun \etal }
    \makebox[\subboxsize]{wSP+M-Net}
    \makebox[\subboxsize]{SP+M+I-Net}
    \makebox[\subboxsize]{$V_{max}$ }
    \makebox[\subboxsize]{Moving-sd. }\\
    \makebox[\subboxsize]{Image}
    \makebox[\subboxsize]{~\cite{hu_iccv2019mask}}
    \makebox[\subboxsize]{~\cite{Cun2020TowardsGS}}
    \makebox[\subboxsize]{(Ours)}
    \makebox[\subboxsize]{(Ours)}
    \makebox[\subboxsize]{Image}
    \makebox[\subboxsize]{Mask}
  
     \caption{\textbf{Comparison of shadow removal methods on the SBU-Vid dataset.}  Qualitative comparison between our methods and other  shadow removal methods: Hu \etal \cite{hu_iccv2019mask} and Cun \etal \cite{Cun2020TowardsGS}. ``wSP+M-Net'' are the  shadow removal results of our patch-based weakly-supervised method. SP+M+I-Net are the shadow removal results of our fully-supervised framework. The last two columns are the $V_{max}$ images and the moving-shadow masks, respectively.
    }
 
    \label{fig:main_video}
\end{figure*}

%% file: main_table.tex
\begin{table}[th]
\centering
\caption{\textbf{Shadow removal results of our fully-supervised method compared to state-of-the-art  shadow removal methods on the adjusted ground truth.}  The metric is MAE (the lower, the better). All methods are trained fully-supervised on the ISTD dataset. Best results are in bold.}
\begin{tabular}{lccc}
\toprule
Methods                    & Shadow& Non-Shadow& All  \\ 
\midrule
Input Image                & 40.2  & 2.6 & 8.5\\ 
\midrule
DeshadowNet \cite{Qu_2017_CVPR}   & 15.9  & 6.0 & 7.6\\ 
Wang \etal \cite{Wang_2018_CVPR}    & 13.4  & 7.7 & 8.7\\ 
Cun \etal~\cite{Cun2020TowardsGS} & 11.4 &7.2 & 7.9\\
Hu \etal~\cite{hu_pami_2019} & 7.6 &3.2 & 3.9\\
\midrule
SP-Net   & 8.4  &3.3 &4.1\\
SP+M-Net & 6.5  &3.1 &3.7\\ 
SP+M+I-Net  & \textbf{6.0}  &\textbf{3.1} &\textbf{3.6}\\ 
\midrule
\textit{SP+M+I-Net (GT-Mask) } & \textit{5.3}  &\textit{2.5} &\textit{2.9}\\ 
\midrule
\multicolumn{4}{c}{Our method with no input shadow-mask}\\
\midrule
SP-Net   & 9.1  &3.1 &4.1\\
SP+M-Net & 8.2  &3.3 &4.0\\ 
SP+M+I-Net  & 6.6  &3.5 &4.0\\ 
\midrule

\end{tabular}
\label{tab:basic}
\end{table}

\setlength{\tabcolsep}{4pt}
\begin{table}[th]
\begin{center}
\caption{\textbf{Shadow removal results of our weakly-supervised method compared to weakly-supervised and priors-based shadow removal methods on the adjusted ISTD testing set \cite{Wang_2018_CVPR}}. }
\label{tab:weakly}
\begin{tabular}{llccc}
\hline\noalign{\smallskip}
Methods   &Training Data                & Shadow& Non-Shadow& All  \\ 
\noalign{\smallskip}
\midrule
\noalign{\smallskip}
Input Image  & -             & 40.2  & 2.6 & 8.5\\ 
\midrule
Yang \etal~\cite{Yang12}   &   -           & 24.7  & 14.4 & 16.0\\ 
Gong \etal~\cite{Gong16}            &-& 13.3  & - & -\\ 
Hu \etal \cite{hu_iccv2019mask}  &Shd. Free (U)        & 12.5  & 3.8 & 5.2\\ \midrule
wSP+M-Net & Shd. Mask & \textbf{9.7}  &\textbf{3.0} &\textbf{4.0}\\
\textit{wSP+M-Net (GT-Mask) } & \textit{Shd. Mask}  & \textit{9.1}  &\textit{2.6} &\textit{3.6}\\ 
\midrule
\end{tabular}
\end{center}
\end{table}
\setlength{\tabcolsep}{1.4pt}

%% file: Sec_7_Conclusion.tex
\section{Conclusions}
In this work, we have presented a novel approach for shadow removal in single images. Our main contribution is to use deep networks as the parameters estimators for an illumination model. 
Our approach has advantages over previous approaches. Compared to early methods using an illumination model for removing shadows, our deep networks can estimate the parameters for the model from a single image accurately and automatically. Compared to deep learning methods that perform shadow removal via an end-to-end mapping, our shadow removal framework outputs images with high quality and much fewer artifacts since we do not use the deep network to output the per-pixel values.
Our model clearly achieves  state-of-the-art shadow removal results on the ISTD dataset and the proposed video dataset, for both fully-supervised \cite{Le-etal-ICCV19} and weakly-supervised \cite{Le_2020_ECCV} settings. 
Our current approach can be extended in a number of ways. A more physically sophisticated illumination model would help the framework to output more realistic images. 
It would also be useful to develop a deep-learning based framework for shadow editing via a physical illumination model.